\documentclass[preprint,10pt]{elsarticle}

\usepackage[utf8]{inputenc}  % UTF-8 input encoding
\usepackage[T1]{fontenc}     % Type1 fonts
\usepackage{lmodern}         % Improved Computer Modern font
\usepackage{microtype}       % Better handling of typo
\usepackage[scaled]{helvet} % Scale Helvetica
\usepackage[english]{babel}  % Hyphenation
\usepackage{graphicx}        % Import graphics
\usepackage[bookmarks=false]{hyperref}            % Better handling of references in PDFs
\usepackage{comment}         % To comment out blocks of text
\biboptions{sort&compress}   % Sort and compress citations

\journal{Environmental Software \& Engineering}

\usepackage{amsmath,amsfonts,amssymb}

\usepackage{subfig}

\usepackage{xcolor}
\usepackage{todonotes}

\usepackage{setspace}
\usepackage{graphicx}
\usepackage{amsmath}
\usepackage{amssymb}
\usepackage{booktabs}
\usepackage[capitalize]{cleveref}
\crefname{section}{Sec.}{Secs.}
\Crefname{section}{Section}{Sections}
\Crefname{table}{Table}{Tables}
\crefname{table}{Tab.}{Tabs.}
\usepackage{algorithm} 
\usepackage{algpseudocode} 
\usepackage[top=1.2in,bottom=1.2in,left=1in, right=1in]{geometry}

\newcommand{\etal}{\textit{et al}.}

\graphicspath{{fig/}}
\begin{document}

\begin{frontmatter}

% Title
\title{Fire Dynamic Vision: Image Segmentation and Tracking for
  Multi-Scale Fire and Plume Behavior}

% Authors and affiliations
  \author{Daryn Sagel\fnref{label1,label2}%
}
\author{Bryan Quaife\fnref{label1,label2}%
}
\fntext[label1]{Department of Scientific Computing, Florida State University, Tallahassee, FL 32306%
}
\fntext[label2]{Geophysical Fluid Dynamics Institute, Florida State University, Tallahassee, FL 32036%
}
\begin{abstract}
The increasing frequency and severity of wildfires highlight the need
for accurate fire and plume spread models. We introduce an approach that
effectively isolates and tracks fire and plume behavior across various
spatial and temporal scales and image types, identifying physical
phenomena in the system and providing insights useful for developing and
validating models. Our method combines image segmentation and graph
theory to delineate fire fronts and plume boundaries. We demonstrate
that the method effectively distinguishes fires and plumes from visually
similar objects. Results demonstrate the successful isolation and
tracking of fire and plume dynamics across various image sources,
ranging from synoptic-scale ($10^{4}$--$10^{5}$~m) satellite images to
sub-microscale ($10^{0}$--$10^{1}$~m) images captured close to the
fire environment. Furthermore, the methodology leverages image
inpainting and spatio-temporal dataset generation for use in
statistical and machine learning models.
\end{abstract}

%\begin{graphicalabstract}
%  \centering
%  \includegraphics[width=\textwidth]{graphicalabstract.pdf}
%\end{graphicalabstract}
%
%\begin{highlights}
%\item FDV is software that processes infrared and visual videos of fire
%  environments.
%\item FDV segments visual videos in RGB-HSV space by leveraging fire
%  environment features.
%\item FDV quantifies and statistically analyzes motion from fire and
%  plume videos.
%\item New insights into fire and plume behavior are discovered.
%\end{highlights}


\end{frontmatter}

%% ---------------------------
%% Introduction
%% ---------------------------
\section{Introduction} %% {{{

The increasing frequency of destructive and costly large-scale wildfires
worldwide, along with the necessity to manage lands using prescribed
fires, underscores an urgent need for a deeper understanding and more
accurate predictive capabilities for fire and plume behavior.  The
complex coupling between the fire and the plume is responsible for the
injection of smoke into the atmosphere and the fire-induced wind
patterns that affect fire behavior.  Understanding these behaviors is
critical for land managers, fire responders, and policy makers who face
an ever-growing wildland urban interface, more intense fires, air
quality regulations, and narrowing prescribed burn windows. The driving
elements of a fire (atmosphere, heat, fuel) are strongly interdependent
and consist of complex turbulent behaviors that vary across physical and
temporal scales. The result is a system that is multifaceted,
multi-physics, and multi-scale in nature and developing models and
numerical methods that are statistically accurate remains a challenge.
As such, a variety of structures and processes must be measured,
modeled, and analyzed for fire environments with scales spanning
sub-microscale (a few meters) to synoptic (tens of kilometers).

Researchers have developed a variety of models that describe fire and
plume dynamics ranging from the landscape scale down to the pyrolysis
and combustion scales~\cite{rego2021fire, linn1997firetec,
parsons2010modeling}. However, models that include a wide range of
scales require large amounts of computational effort. Alternatively,
simplified models~\cite{achtemeier2011modeling, finney2006overview,
linn2020quic, kolaitis2023comparative, robinson2023effect} can be
applied in operational settings and complement the expertise of fire
practitioners. Regardless of the model used, it must be validated and
calibrated with data collected in the field~\cite{andrews2018rothermel}.

Instruments that measure wind speed, temperature, fuels, ember spread,
and more are often expensive and record sparse measurements at only one
or two locations. Additionally, equipment must be protected from the
fire, often making it difficult or impossible to measure fine-scale
features in the combusting domain. For example, a sonic anemometer
compiles wind speed data at a location away from the fire, and the
measurements are extrapolated to assign a potentially inaccurate wind
speed at the fire's location.

Alternatively, cameras are an excellent option for capturing features of
fires and plumes, and can record data at a safe distance from the fire.
For example, fires are commonly detected by combining machine learning
tools~\cite{bu2019intelligent, ji2024machine} with images captured by
cameras mounted on fixed structures, unmanned aerial vehicles (UAVs),
and satellites. However, the detection of fire in an image does not
necessarily translate to capturing detailed fire and plume dynamics. To
capture these dynamics, researchers often use computer vision tools and
black-box algorithms designed in other scientific domains, but such
methods are usually ill-suited for studying turbulent motions in fire
environments.

{\bf We present Fire Dynamic Vision (FDV), a multi-scale and multi-view tool
that segments fires and plumes in visual and infrared videos and
statistically interprets their dynamics.}

%% ---------------------------
%% Related Work
%% ---------------------------
\subsection{Related Work} %% {{{
\label{sec:related}

This section reviews the literature of key components of FDV: image
segmentation, cluster analysis, and fire and plume tracking. FDV
incorporates methodologies that are less prevalent in fire research but
are widely used in classical computer vision and image processing. This
section also provides a brief overview of the related problem of fire
detection.

%%%%%%%%%%%%%%%%%%%%%%%%%%%%%%%%%%%%%%%%%%%%%%%%%%%%%%
\subsubsection{Image Segmentation and Cluster Analysis}
\label{sec:relatedseg}

Relying solely on RGB segmentation has long been considered inadequate
for many computer vision applications~\cite{maheswari2016review}.
However, the HSV color space is well-documented as a successful option
for segmentation~\cite{bora2015comparing, hema2020interactive,
huang2007segmentation, li2014fire, zhang2008new}, particularly when
differentiating objects with similar RGB colors. Researchers have
experienced similar success using hybrid RGB-HSV color
segmentation~\cite{mythili2012color}, with applications ranging from
medical imaging to satellite images~\cite{ganesan2014assessment,
hassan2017color}. Within fire studies, additional common color space
choices include CIE-L*a*b*~\cite{truong2012fire} and
YCbCr~\cite{chino2015bowfire, celik2008computer, chen2021novel}.
 
Image segmentation is frequently followed by cluster
analysis~\cite{naik2014review, sathya2011image}. Popular methods include
$k$-means~\cite{coleman1979image} and fuzzy
$c$-means~\cite{chuang2006fuzzy}, which are regularly applied to
satellite images~\cite{abburu2015satellite, hamada2019multi}. Another
popular method is density-based spatial clustering of applications with
noise (DBSCAN)~\cite{khan2014dbscan}, which has been shown to outperform
$k$-means when applied to noisy
datasets~\cite{chakraborty2014performance}.

%%%%%%%%%%%%%%%%%%%%%%%%%%%%%%%%%%%%%%%%%%%%%%%%%%%%%%
\subsubsection{Video-Based Tracking}
\label{sec:relatedtrack}

Tracking the boundary of a fire or plume is important for both
researchers and practitioners. Fire and plume dynamics are often tracked
with particle image velocimetry (PIV)~\cite{coen2004infrared,
desai2022investigating, morandini2012feasibility} or thermal image
velocimetry (TIV)~\cite{katurji2021turbulent, schumacherrate} applied to
visual or infrared videos. PIV and TIV are optical flow methods that
track evolving structures by relating points of interest and temperature
regions, respectively, between successive frames. Videos may be taken in
a laboratory~\cite{martinez2006laboratory, zhou2003thermal} or a
field~\cite{paugam2012use, stow2014measuring} setting. However, there
are concerns among researchers and model developers about the extent to
which laboratory dynamics translate to the dynamics of a fire in natural
environments~\cite{raposo2018analysis}.

%%%%%%%%%%%%%%%%%%%%%%%%%%%%%%%%%%%%%%%%%%%%%%%%%%%%%%
\subsubsection{Video-Based Detection}
\label{sec:detection}

In recent years, the field of machine learning has seen a significant
increase in the use of visual video for fire
detection~\cite{ccetin2013video, liu2003review, zhang2009image}. This
surge in interest is partly attributable to the availability of large
satellite image datasets~\cite{cardil2023characterizing,
ganesan2016comparative, wooster2021satellite, ray2023characterization,
roy1999multi} and the distinctive visual appearance of flaming regions
and smoke plumes. However, the relatively low spatial resolution of
satellite images poses challenges in characterizing dynamics and
detecting small fires~\cite{wu2021forest}. Consequently, certain
vulnerable areas have strategically deployed fixed cameras to provide
continuous monitoring and alerts regarding local fire
activity~\cite{seebamrungsat2014fire, wang2009early}. This has led to
the creation of standard video-based fire recognition methods at smaller
spatial scales~\cite{benjamin2016extraction}, though many of these
methods are not designed for the fine-scale segmentation and analysis
needed to thoroughly study the structures and dynamics. Moreover, while
the study of fire behavior heavily relies on infrared
images~\cite{johnston2018flame}, few methods in fire detection from
ground-based imagery incorporate infrared
images~\cite{moumgiakmas2021computer} and even fewer combine both visual
and infrared video.

%%%%%%%%%%%%%%%%%%%%%%%%%%%%%%%%%%%%%%%%%%%%%%%%%%%%%%%%%%%%
\subsection{Contributions}

\begin{figure*}[ht!]
  \begin{center}
    \includegraphics[width=0.9\linewidth,trim=0cm 1.5cm 0cm 1cm,clip=true]{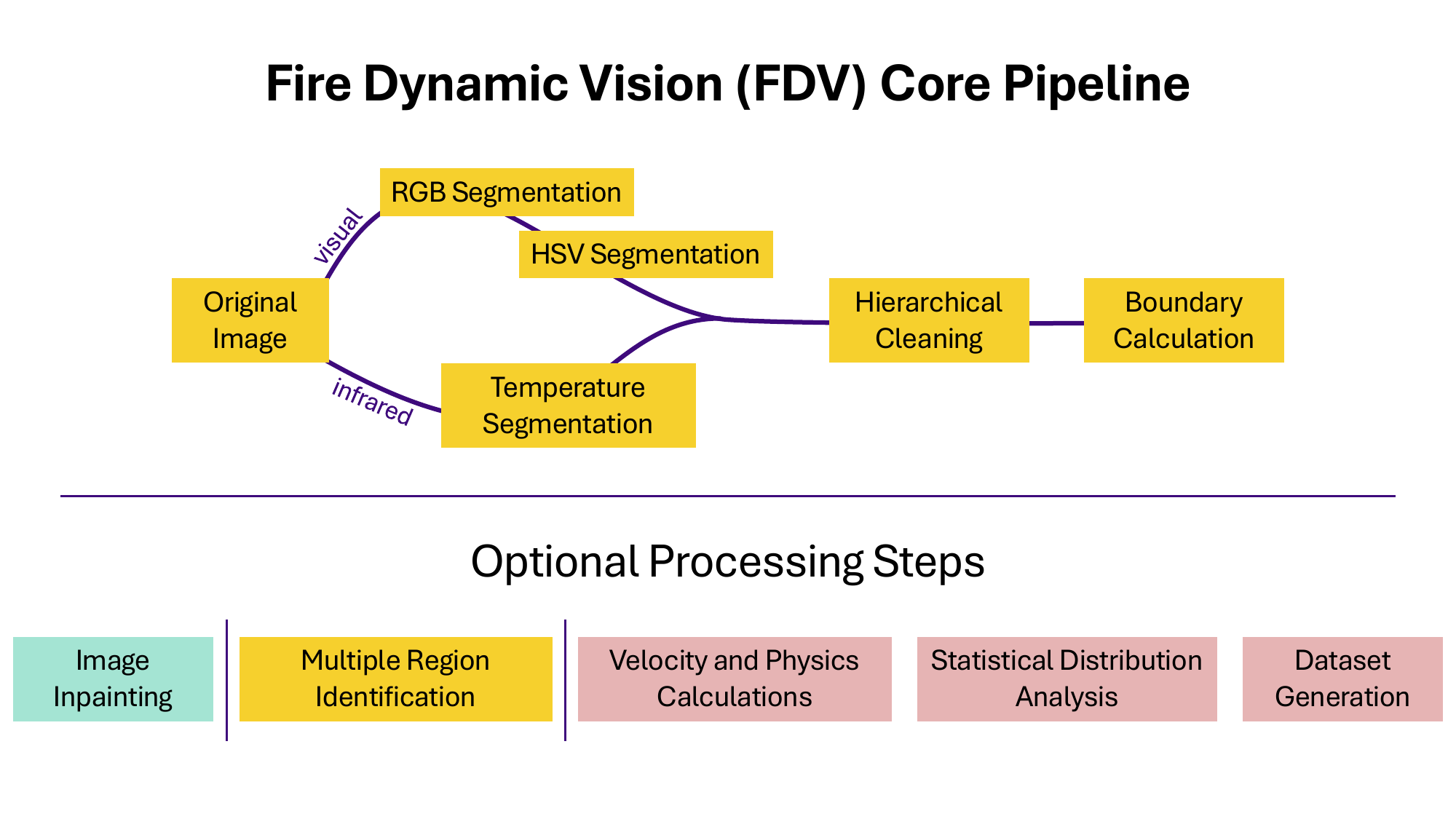}
  \end{center}
   \caption{\label{fig:overview} FDV uses segmentation with
   spatial cluster analysis to locate flaming regions, burned regions,
   and plume boundaries. These boundaries are optionally tracked between
   frames, yielding important insights into the evolution of fire fronts
   and plumes in diverse situations. FDV also supports image inpainting
   for pre-processing, multiple region identification during processing,
   and several post-processing options including statistical analysis of
   the calculated distributions and automatic dataset generation.
   Pre-processing is denoted with a green box, in-process steps are
   denoted with gold, and post-processing steps are denoted with red.}
\end{figure*}

FDV builds on our
previous work~\cite{sagel2021fine} by incorporating a novel combination
of computer vision techniques, thereby expanding the range of fire
configurations and video types that can be effectively analyzed. In
particular, FDV: (1) extends the scope of segmentation and tracking by
including both visual and infrared images; (2) allows for segmentation
and tracking of both fires and plumes, whereas previously the focus was
solely on fires; and (3) enables the processing of videos captured from
various viewpoints, in contrast to our earlier approach that only
accommodated top-down perspectives. FDV combines RGB (Red, Green, Blue)
and HSV (Hue, Saturation, Value) image segmentation to leverage key
visual assumptions of fire environments, and includes a spatial cluster
analysis step to process videos with multiple fronts. FDV is a valuable
and versatile tool for studying fire and plume behavior. It provides a
means of validating and improving existing methods through observational
data, thus complementing various models, including those for fire
detection and spread prediction. Additionally, FDV supports the
development of new physical and data-driven models that aid in our
understanding of fire and plume dynamics. FDV's comprehensive pipeline
is illustrated in Figure~\ref{fig:overview}.

To demonstrate the utility of FDV, we apply it to a selection of videos
capturing fire and plume spread with resolutions ranging from
millimeters per pixel to several hundred meters per pixel. Additionally,
we analyze fire and plume spread from different camera angles and employ
Markov Chain Monte Carlo (MCMC) to statistically characterize the
dynamics of fire and plume spread and enhance our understanding of the
underlying processes. Finally, we introduce dataset generation
functionality that is useful for training machine learning and
statistical models.

The remainder of the paper is outlined as follows:
Section~\ref{sec:methodology} details the methodology implemented within
FDV, followed by practical applications and an analysis of the results
in Section~\ref{sec:experiments}. Finally, concluding remarks are
presented in Section~\ref{sec:conclusion}.

%% ---------------------------
%% Methodology
%% ---------------------------
\section{Methodology} %% {{{
\label{sec:methodology}

FDV extends our previous work~\cite{sagel2021fine} to allow for
segmentation, tracking, and analysis using both visual and infrared
images of fire environments from different angles, as well as the
ability to track multiple fire fronts and plumes. FDV segments visual
images using RGB and HSV color spaces (Section~\ref{sec:seg}), while
infrared images continue to be segmented according to
temperature~\cite{sagel2021fine}. Then, spatial cluster analysis
classifies pixels as belonging to the fire region, burned region, or
plume and a boundary is calculated for each region of interest
(Section~\ref{sec:clus}). Finally, displacements are calculated between
boundaries in successive frames and a statistical framework for
interpreting this motion is introduced (Section~\ref{sec:trac}). An
additional feature, mostly of interest to fire behavior modelers,
converts the segmented regions into a spatial fire spread dataset
(Section~\ref{sec:dset} and Section~\ref{sec:image}). Pseudocode
summarizing the pipeline is provided in Section~\ref{sec:pipe}.

FDV is developed with a primary focus on usability, regardless of
available resources. It operates efficiently even with cell phone videos
and a standard laptop, thus eliminating the need for high-end cameras or
supercomputers. Additionally, FDV does not rely on machine learning,
removing the requirement for labeled datasets. It is our experience that
large, appropriately labeled, and standardized training sets of
high-resolution fire images from different angles, scales, and
environments are currently unavailable. Although many fire detection
studies use machine learning to detect fires and plumes, as discussed in
Section~\ref{sec:detection}, these datasets primarily consist of
satellite and UAV images, thereby restricting the scales at which a
parameterization can reliably predict fire and plume spread.

%% ---------------------------
%% Segmentation
%% ---------------------------
\subsection{Segmentation} %% {{{
\label{sec:seg}

\begin{figure*}[ht]
  \begin{center}
    \includegraphics[width=0.9\linewidth,trim=0cm 1cm 0cm 0cm,clip=true]{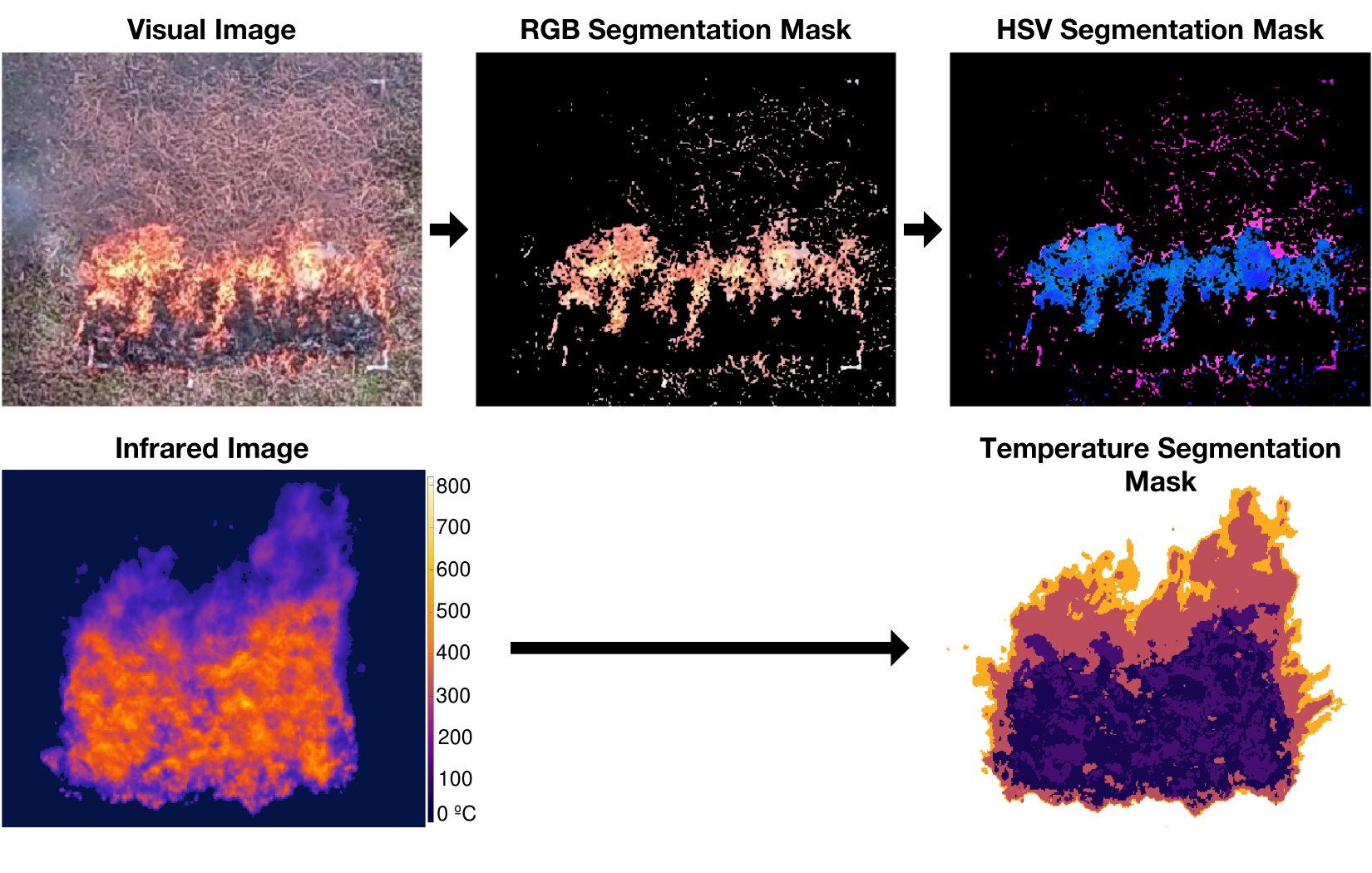}
  \end{center}
  \caption{Sample image segmentation using FDV for a visual image ({\em top row}) and an infrared image ({\em bottom row}). Visual image segmentation
  involves two steps: RGB segmentation ({\em top middle}), which captures the
  fire and surrounding pine straw, and HSV segmentation ({\em top right}),
  where the fire (blue) and pine straw (pink) are distinct due to their
  color saturation differences. Infrared image segmentation is a single
  step thresholding step resulting in a temperature-based mask ({\em bottom
  right}) that shows the burned and cooled region (dark purple), burning
  region (red), and pre-heated region (yellow). Note that the visual and
  infrared frames are from different timesteps, chosen separately for
  visual clarity.}
\label{fig:overview2}
\end{figure*}

\begin{figure*}[h!]
\begin{center}
\includegraphics[width=0.9\linewidth]{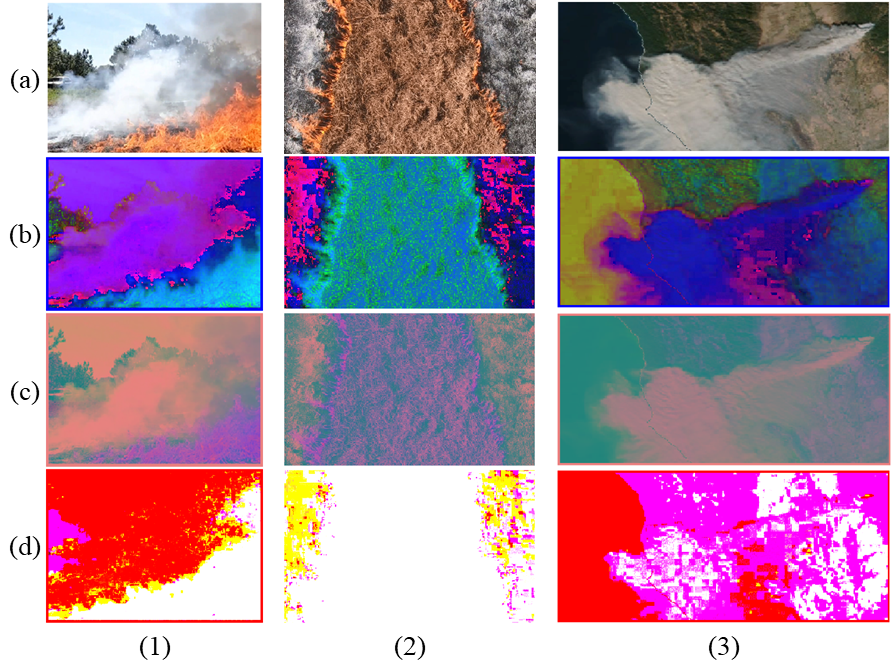}
\end{center}
  \caption{\label{fig:colors} Visualization of the fire, plume, and
  background in (a) RGB, (b) HSV, (c) YCbCr, and (d) CIE-L*a*b* color
  spaces for a (1) sub-microscale side-view, (2) sub-microscale overhead
  view, and (3) synoptic-scale overhead view. Segmentation in YCbCr
  color space offers fire, plume, and background distinction similar to
  HSV, but requires additional computation time to convert from RGB
  color space. CIE-L*a*b* color space segmentation is a visibly poor
  choice for this application.}
\end{figure*}

Regions of interest in fire videos include the flaming area, burned
area, and plume. In visual images of these environments, regions of
interest are often colored similarly to their surroundings. This
similarity is especially evident in the southeastern United States since
flames and pine straw share similar RGB values, as shown in
Figure~\ref{fig:overview2}, and RGB segmentation captures a significant
amount of the pine straw fuel bed along with the fire. However, fire has
distinct HSV values compared to its surroundings due to its brighter and
more saturated appearance. As such, incorporating HSV color segmentation
alongside RGB segmentation helps distinguish fire from visually similar
objects, as depicted in Figure~\ref{fig:overview2}. Plume segmentation
faces similar challenges, often requiring a highly specific set of RGB
ranges that can be challenging to pinpoint due to color variations
caused by fuel type and smoke density, as well as plumes' similarity to
clouds. As such, FDV employs a hybrid approach using both RGB and HSV
color spaces that offers a clear distinction between the fire or plume
and its background. FDV leverages this combination to automatically
segment images with pixel value thresholding, bypassing the need for
more complex thresholding techniques~\cite{raju2012image}. As discussed
in Section~\ref{sec:relatedseg}, other popular color space choices
within fire science and general image processing include CIE-L*a*b* and
YCbCr. Figure~\ref{fig:colors} compares these color spaces applied to
fire environments. It is apparent from this
visualization that RGB and HSV best fit our processing pipeline and
applications.

Fire tracking algorithms typically omit a segmentation step, which
generally does not pose issues for infrared data due to mid- and
long-wave infrared sensors recording observations with minimal
distortion from smoke. However, visual cameras cannot penetrate smoke,
leading to occlusions in visual fire videos. For example, this is
problematic for analyzing overhead videos, like in
Figure~\ref{fig:overview2}, because rising and expanding smoke appears
to create unphysical diverging wind velocity vectors in two-dimensional
projections along the horizontal plane. By including image segmentation,
FDV addresses this common issue and reduces inaccuracies associated with
fire tracking from image data.

%% ---------------------------
%% Clustering
%% ---------------------------
\subsection{Boundary Calculation and Cluster Analysis} %% {{{
\label{sec:clus}

Segmentation results in a binary mask that often includes points beyond
those we wish to track. Therefore, the segmented images are cleaned
before identifying the boundaries of the main bodies of the fire or
plume. A simple single-level cleaning approach eliminates points with
fewer than a user-specified number of nonzero neighboring
pixels~\cite{sagel2021fine}. In our experience, the cleaned and isolated
regions are highly sensitive to this tolerance; if the tolerance is too
small then important features are lost, and if it is too large then
unimportant points are retained. To combat this sensitivity, FDV employs
a coarse-to-fine hierarchical search that identifies both large and
small areas in the image to retain or exclude.

FDV's cleaning approach starts with a large radius and neighbor
threshold to capture large structures and progressively refines to a
small radius and neighbor threshold to capture rogue points or small
point clusters. This hierarchical method is superior to the single-level
cleaning approach, which is impractical due to the significant size
variations of the regions that need to be excluded. Although the
hierarchical search is computationally more expensive than the
single-level search, this additional cost is amortized---because the
removal is more effective, the cost of other algorithmic steps with
higher computational complexity are significantly reduced.
%Sample results of hierarchical cleaning are shown for the fire and
%plume in the third row of Figure~\ref{fig:overview4}.

After cleaning, DBSCAN, a spatial cluster analysis method, identifies
distinct regions in each image, such as multiple fire fronts or plume
structures.  Unlike the common choice of $k$-means, DBSCAN does not
require a user-specified number of clusters a priori, making it more
suitable for segmenting images with an unknown number of regions of
interest.  DBSCAN's two tuning parameters are selected to accommodate
physical constraints on fire and plume dynamics, and selected values are
reported in Section~\ref{sec:future}. Additionally, since DBSCAN is
designed to handle noisy datasets, it is ideal for videos of fire
environments that inherently contain noise. We experimented with other
cluster analysis methods, such as fuzzy $c$-means and spectral
clustering~\cite{chen2020dbscan, murugesan2019benchmarking,
schubert2018relationship}, and found DBSCAN to be the fastest and most
robust choice for FDV's intended applications. Specifically, DBSCAN
excels at distinguishing regions separated by curves, whereas other
tested cluster analysis algorithms function best with linear separations
or separations with small curvatures.

\begin{figure*}[h]
\begin{center}
\includegraphics[width=0.3\linewidth]{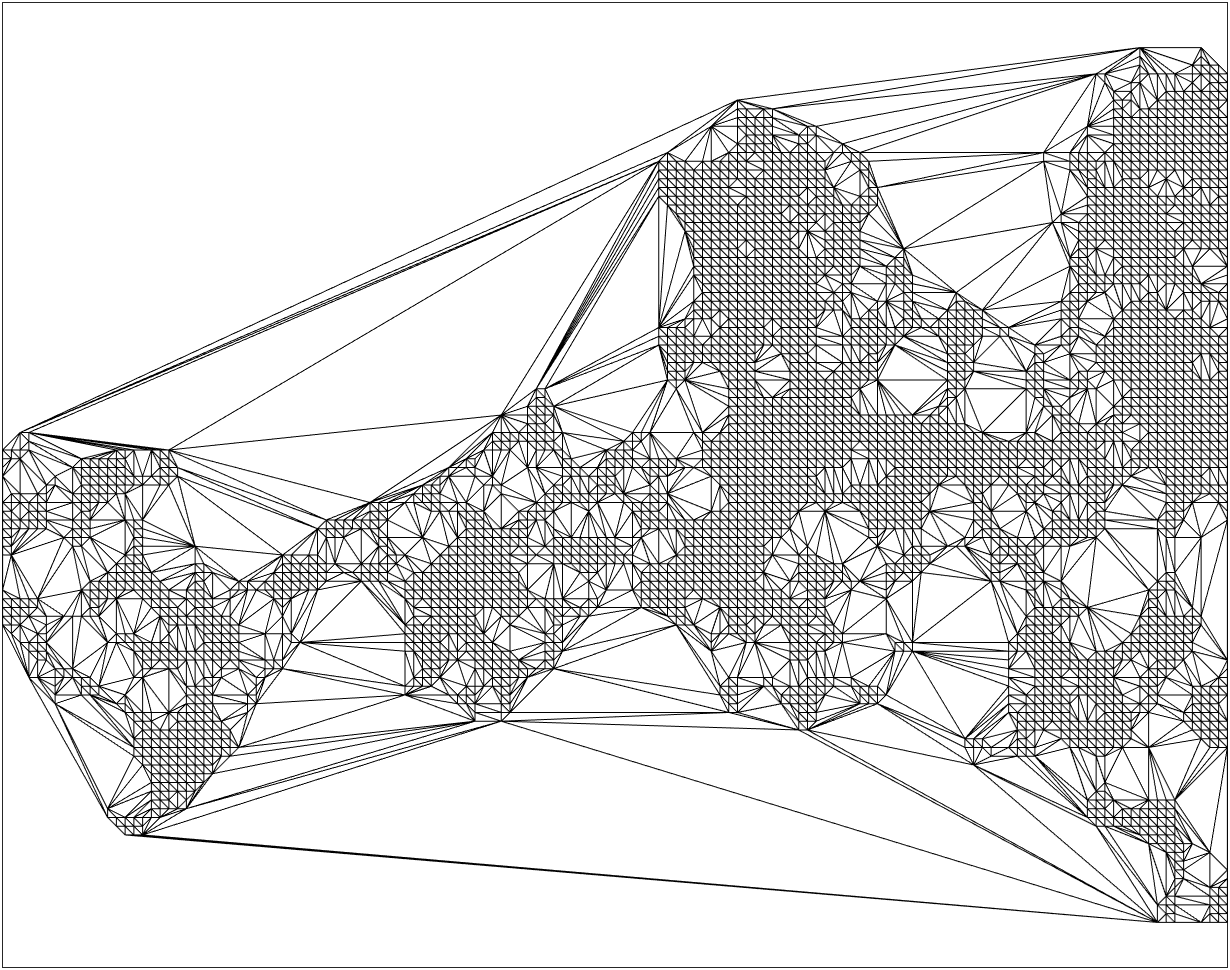}
\includegraphics[width=0.3\linewidth]{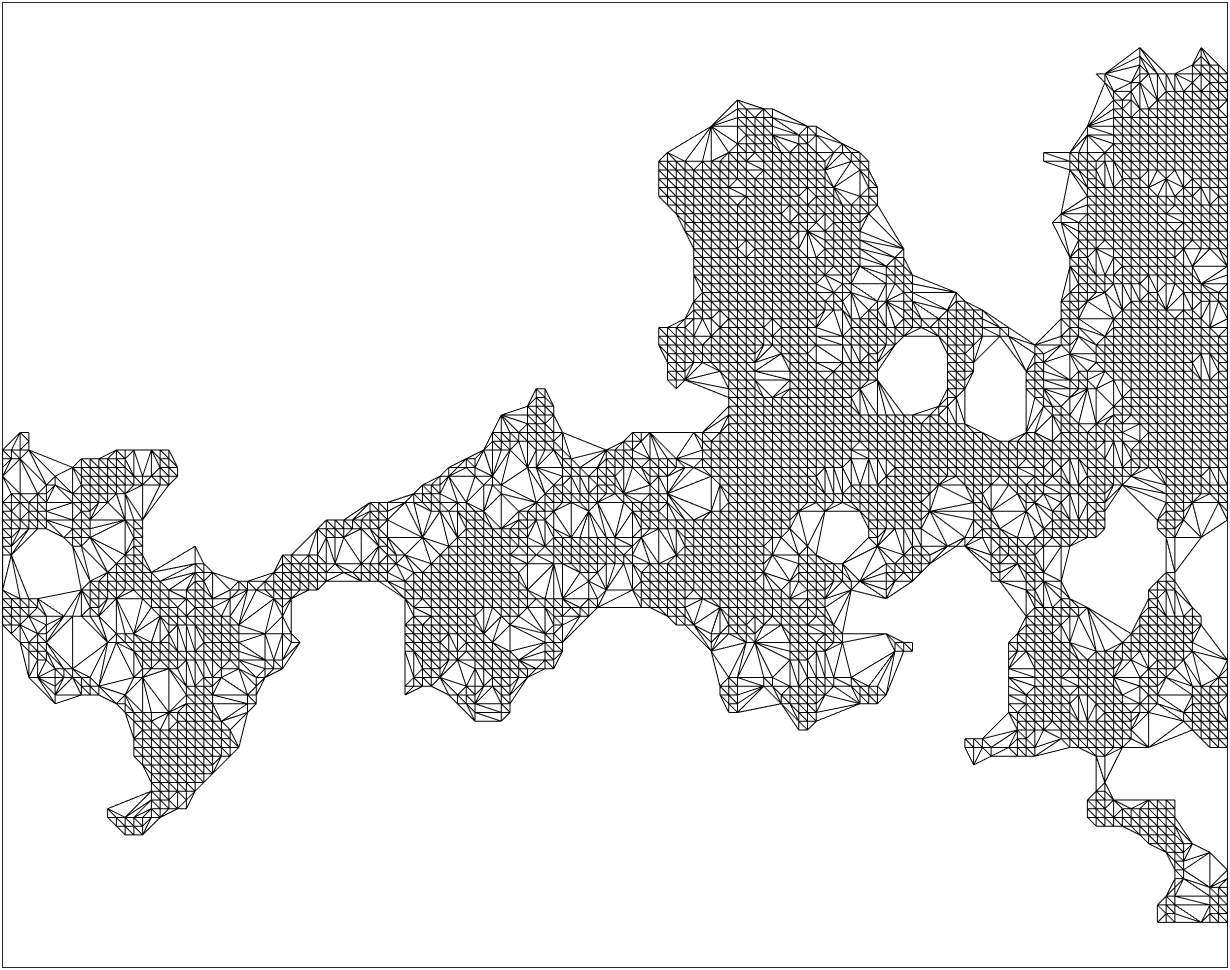}
\includegraphics[width=0.3\linewidth]{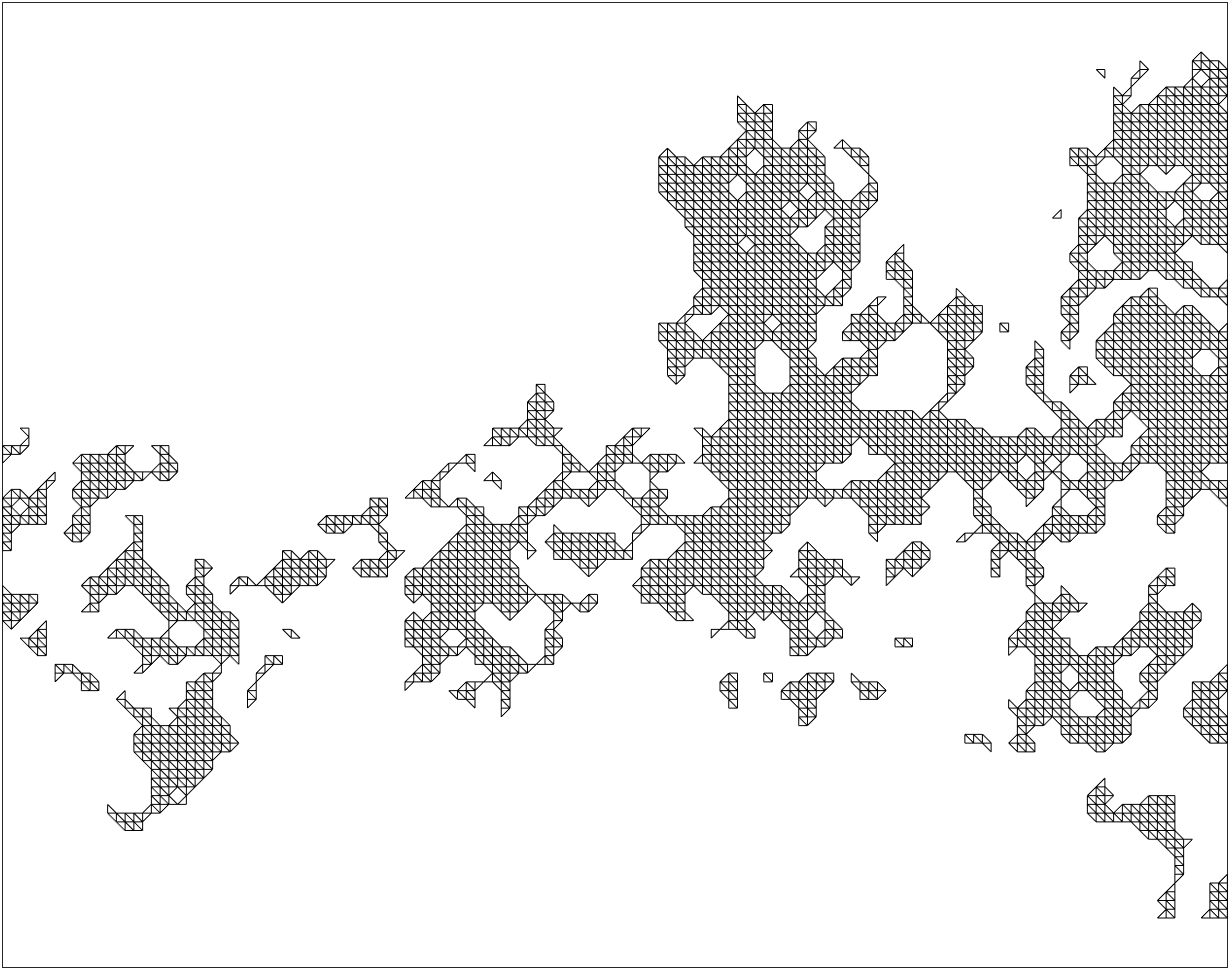} \\
\includegraphics[width=0.3\linewidth,trim=0.1cm 2cm 0.1cm 4cm,clip=true]{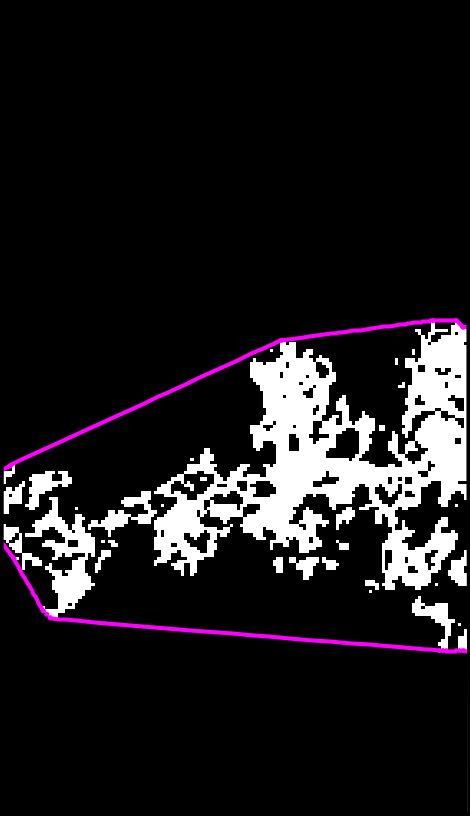}
\includegraphics[width=0.3\linewidth,trim=0.1cm 2cm 0.1cm 4cm,clip=true]{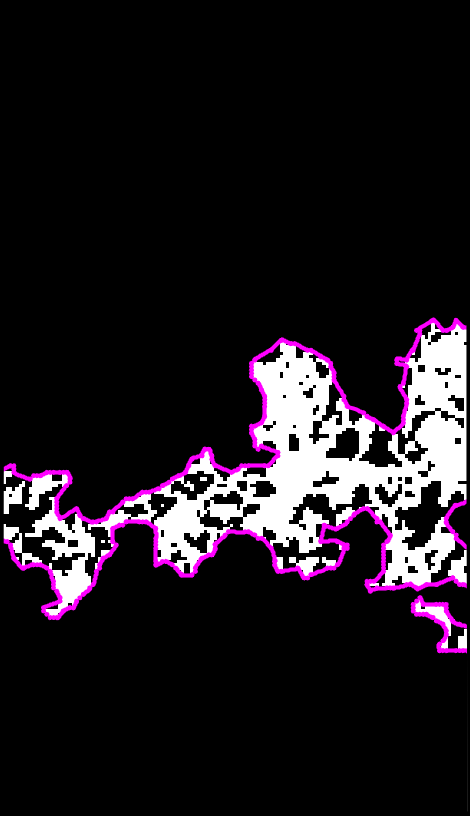}
\includegraphics[width=0.3\linewidth,trim=0.1cm 2cm 0.1cm 4cm,clip=true]{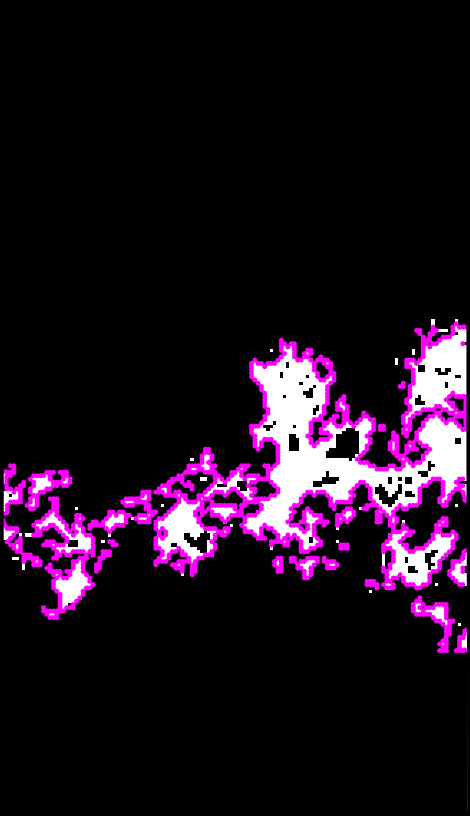}
\end{center}
   \caption{The $\alpha$-shape ({\em top row}) and resulting boundary ({\em bottom
   row}) for sample values of the radius $\alpha$, applied to a cleaned
   segmentation mask from an overhead image of fire spread. When
   $\alpha=0$ ({\em left}), the result is the convex hull. In this paper, we
   use $\alpha=1/3$ ({\em middle}). While $\alpha=5/4$ ({\em right}) provides more
   detail, this level of detail increases the impact of noise and
   turbulence on the dataset.}
\label{fig:alphagraphic}
\end{figure*}

The points within each cleaned region form a point cloud that requires a
defined boundary. We utilize the $\alpha$-shape, which extends the
concept of the convex hull, with the parameter $\alpha$ governing the
level of concavity in the shape. This shape is constructed using edges
derived from the Delaunay triangulation~\cite{edelsbrunner2010alpha} of
the point cloud. FDV combines an efficient Delaunay triangulation
algorithm with criteria to determine each edge's inclusion in the
$\alpha$-shape's boundary, thus defining a single-pixel thick boundary
around each region of interest. Large $\alpha$ values result in more
detailed shapes, small $\alpha$ values result in less detailed shapes,
and the limiting value $\alpha=0$ results in the convex hull. After
experimenting with several images and values, we find that $\alpha =
1/3$ results in reasonable boundaries, and this value is used throughout
the remainder of the paper. The $\alpha$-shape method also identifies
and outlines regions that contain gaps. Such gaps are common in videos
of plumes, ring fires, merging fire fronts, or in areas where tree
canopies block portions of the capture area.
Figure~\ref{fig:alphagraphic} illustrates how the choice of $\alpha$
determines the resulting boundary. 

\begin{figure*}[!ht]
\begin{center}
\includegraphics[width=0.25\linewidth,trim=0cm 2cm 0cm 3cm,clip=true]{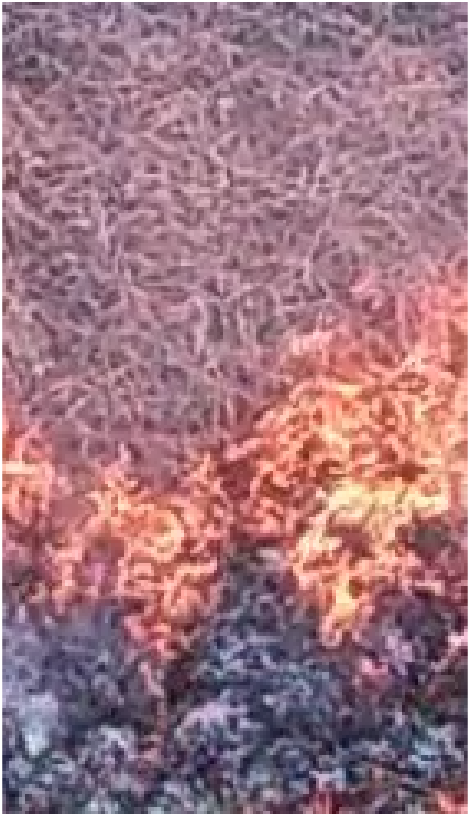}
\includegraphics[width=0.67\linewidth,trim=8cm 0cm 1cm 0cm,clip=true]{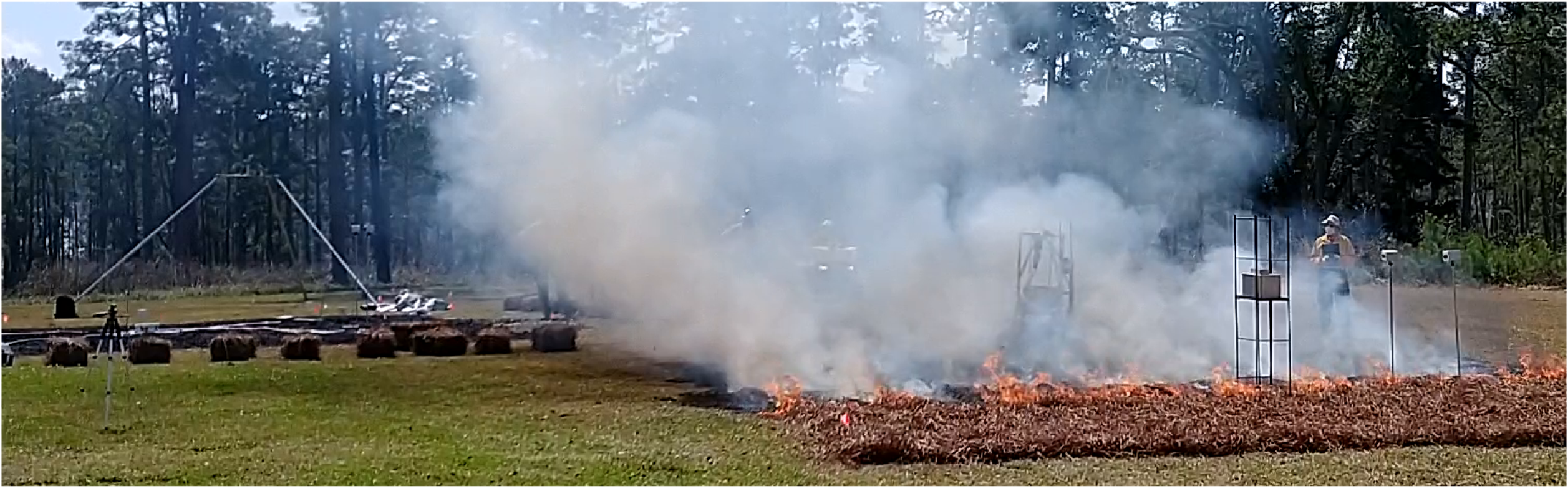} \\
\includegraphics[width=0.25\linewidth,trim=0cm 2cm 0cm 3cm,clip=true]{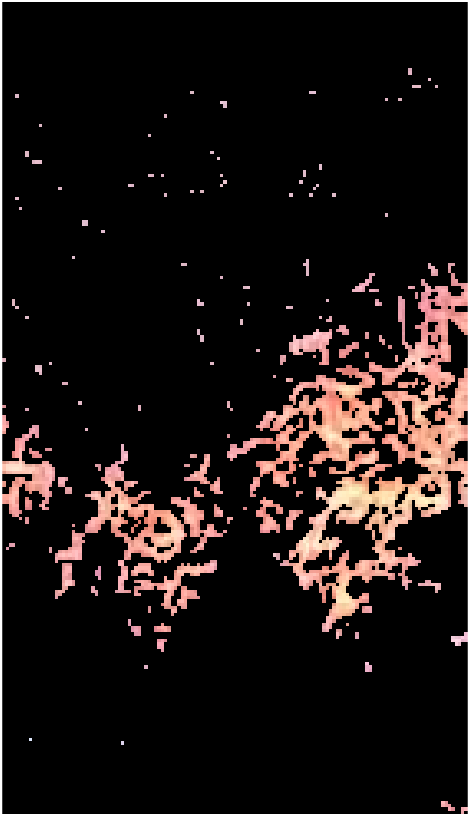}
\includegraphics[width=0.67\linewidth,trim=8cm 0cm 1cm 0cm,clip=true]{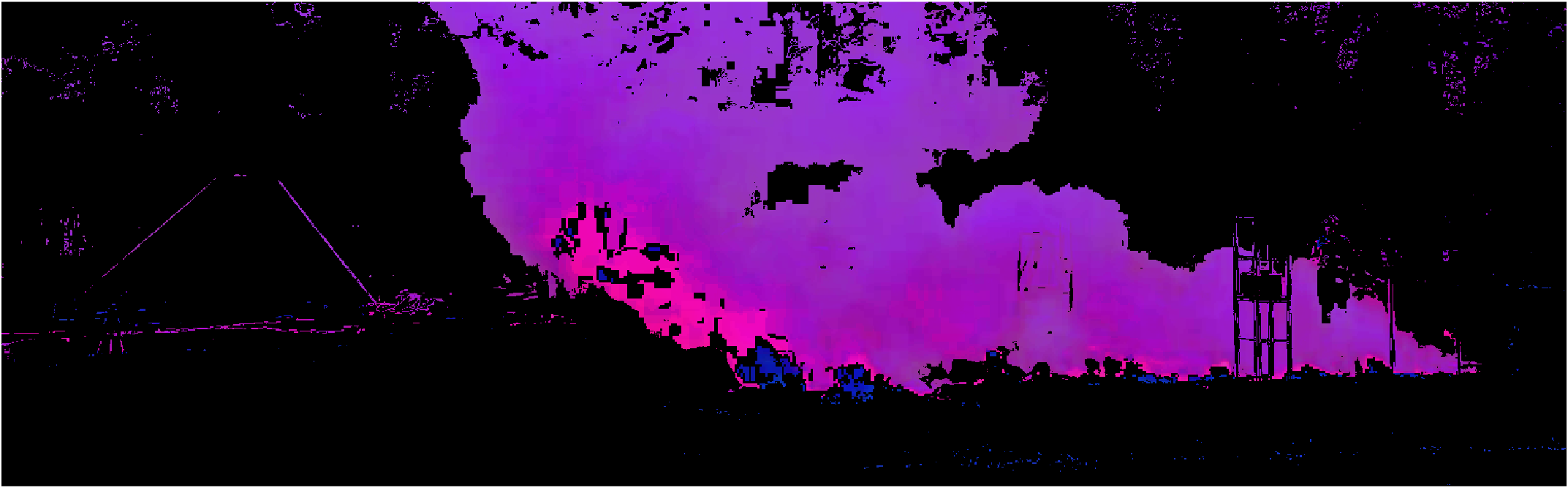} \\
\includegraphics[width=0.25\linewidth,trim=0cm 2cm 0cm 3cm,clip=true]{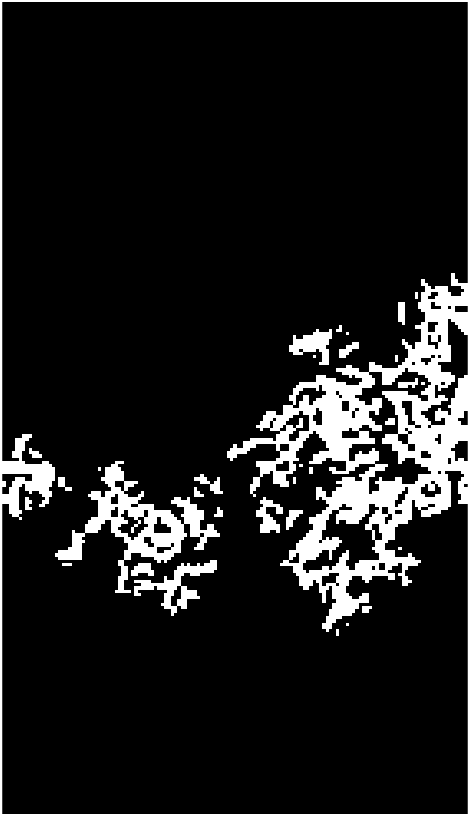}
\includegraphics[width=0.67\linewidth,trim=8cm 0cm 1cm 0cm,clip=true]{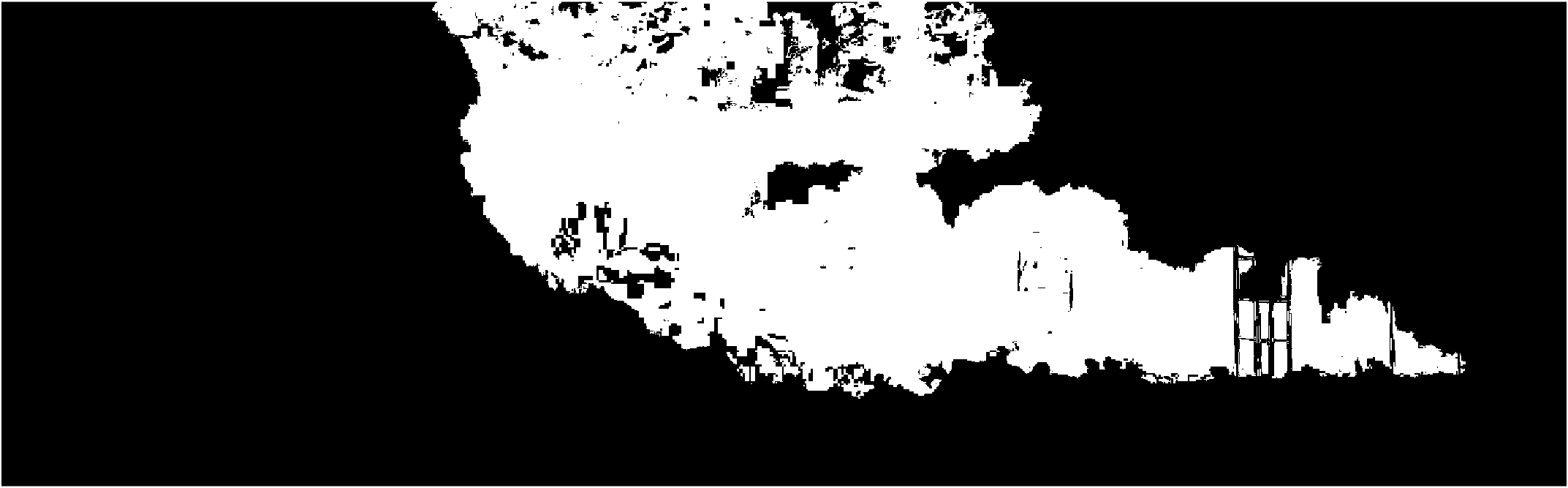} \\
\includegraphics[width=0.25\linewidth,trim=0cm 2cm 0cm 3cm,clip=true]{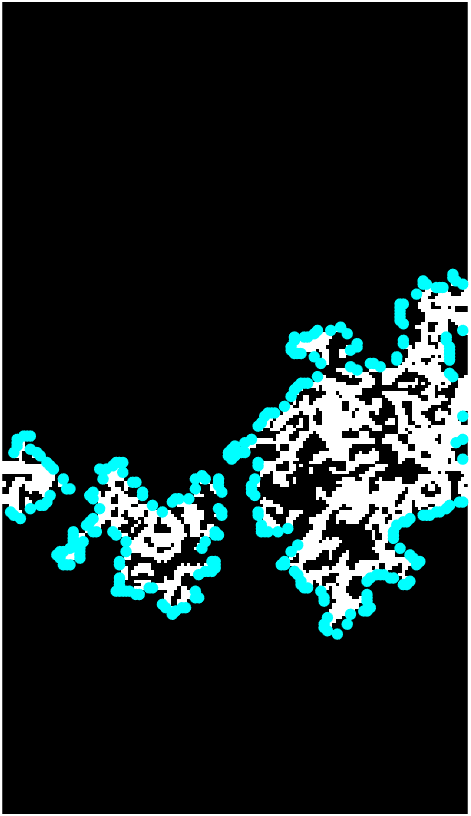}
\includegraphics[width=0.67\linewidth,trim=8cm 0cm 1cm 0cm,clip=true]{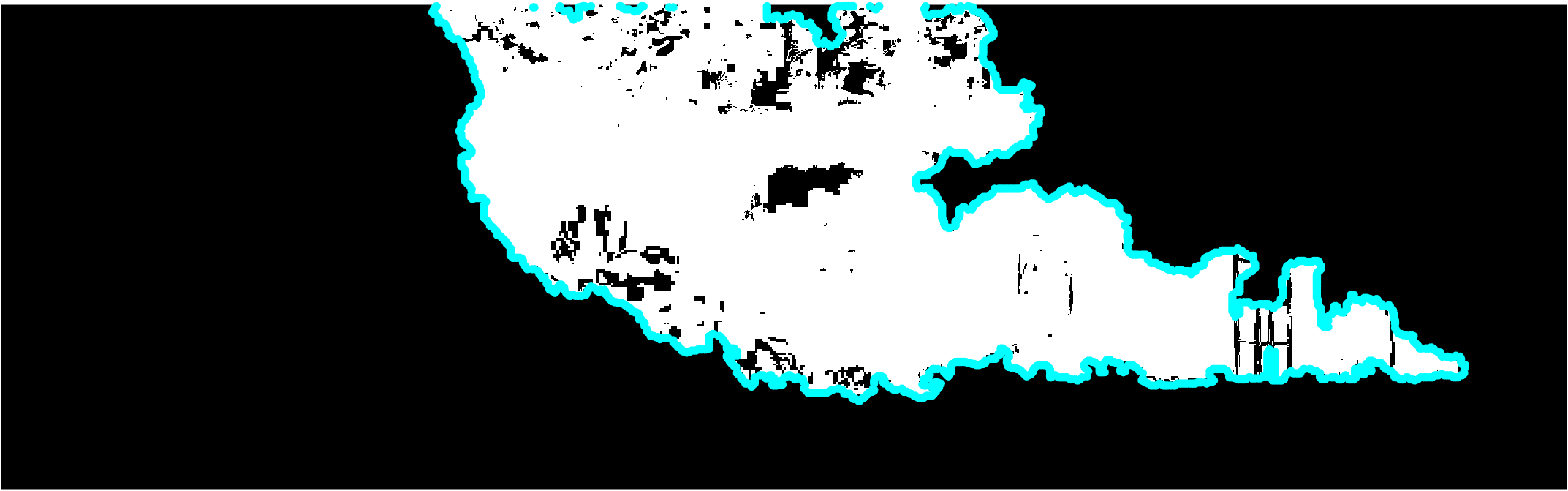}
\end{center}
  \caption{FDV pipeline applied to a visual overhead image of fire ({\em left
  column}) and a visual side-view image of a plume ({\em right column}). The
  original image ({\em first row}) undergoes thresholding to produce a binary
  segmentation mask ({\em second row}). This mask is then cleaned ({\em third row}),
  and the $\alpha$-shape is constructed using the remaining points. The
  $\alpha$-shape determines the boundary points for the region of
  interest, depicted in blue ({\em fourth row}).}
\label{fig:overview4}
\end{figure*}

Figure~\ref{fig:overview4} demonstrates two sample applications of the
FDV pipeline, beginning with the original images (first row), followed
by the segmentation masks (second row), then the cleaned masks (third
row), and ending with the boundaries (fourth row).

%% ---------------------------
%% Tracking
%% ---------------------------
\subsection{Tracking and Statistical Analysis} %% {{{
\label{sec:trac}

After images are segmented and a boundary is defined, FDV calculates
localized displacements of boundary points between successive frames.
Our previous work calculates displacements with a modified assignment
problem~\cite{sagel2021fine}, which requires the dynamics to be largely
concentrated in one particular direction, such as in a head fire. This
method does not accommodate more general dynamics, like those found in
ring fires and plume motions. Additionally, the assignment problem
enforces one-to-one matching, which is not physical; some fuel cells
extinguish while others ignite multiple neighbors, and smoke parcels may
dissipate in some areas while concentrating in others. We replace the
modified assignment problem with greedy nearest neighbor matching, which
permits some points to have no matches while others can have multiple
matches~\cite{austin2014comparison,cormen2022introduction}. A more
thorough comparison between matching methods---including how each method
addresses diverging versus non-diverging flow---will be considered in
future work. In this paper, greedy nearest neighbor matching is applied
to the boundary points obtained through $\alpha$-shape calculation,
resulting in displacement vectors that are easily converted to velocity
vectors by multiplying by the camera's frame rate.

Once velocity vectors are calculated between all pairs of successive
frames, distributions of these vectors are constructed. They often
resemble well-known distributions or their special cases, and our
previous work matched the statistical moments of the data to those of
the distributions~\cite{sagel2021fine}. This approach does not provide
error bounds on the distributions, and we find that
MCMC~\cite{punskaya2002bayesian} produces better fitting results and
provides much-needed error bounds.Results quantifying the increase in
accuracy using MCMC over moment matching are provided in
Section~\ref{sec:small_xy}.

%% ---------------------------
%% Dataset Generation
%% ---------------------------
\subsection{Dataset Post-Processing} %% {{{
\label{sec:dset}

Our previous work calculates datasets of quantities of interest without
associating each data point with specific values in
space-time~\cite{sagel2021fine}. Many applications require these
quantities on a rasterized grid for analysis, such as examining spatial
correlations. Standardized, regular data can also be used to train
machine learning models, such as deep neural
networks~\cite{ton-qua2024}. For example, machine learning models for
fire and plume spread need spatially-represented datasets showing the
positions of actively burning regions or smoke over time in diverse
scenarios. These models, as with many other machine learning
applications, are often trained on simulated data because a sufficient
amount of measured data is not
available~\cite{chouhan2023surrogate,chouhan4853898surrogate} (see
discussion in Section~\ref{sec:seg}).

Rather than relying on simulated data constrained by model assumptions,
FDV provides an alternative route to generate real-world spatio-temporal
information tailored to user-specified regions and quantities of
interest. Datasets are automatically generated and formatted based on
selected attributes, such as fire and plume boundaries at every
timestep, velocity at each point along the fire or plume for each
timestep, the distribution fits of calculated velocities, or
physics-based calculations.

%% ---------------------------
%% Image Inpainting
%% ---------------------------
\subsection{Image Inpainting}
\label{sec:image}

We expand FDV's dataset generation capabilities by optionally
incorporating image inpainting~\cite{elharrouss2020image,
guillemot2013image} to address occlusions, such as gaps in segmentation
results caused by trees in the frame. This enables researchers requiring
spatially continuous datasets to utilize a wider range of video sources
without additional limitations imposed by features of available videos.
Moreover, inpainting helps reduce false velocity vector calculations at
the boundaries of occlusions that intersect areas of interest, such as
when a swaying tree partially overlaps with the area of fire spread in a
video.

The inpainting algorithm, developed by
Bertalmio~\etal~\cite{bertalmio2001navier}, is based on fluid dynamics
principles. We choose this algorithm for two main reasons: (1) fluid
equations are commonly used to model the shape and behavior of fire
systems, and (2) it produces similar results to the popular Fast
Marching Method algorithm developed by Telea~\etal~\cite{telea2004image}
but in less time. Bertalmio's algorithm minimizes variance to fill colors, a
method rooted in physics that FDV integrates through its displacement
matching calculations.

\begin{figure*}[h!]
\begin{center}
\includegraphics[width=0.4\linewidth]{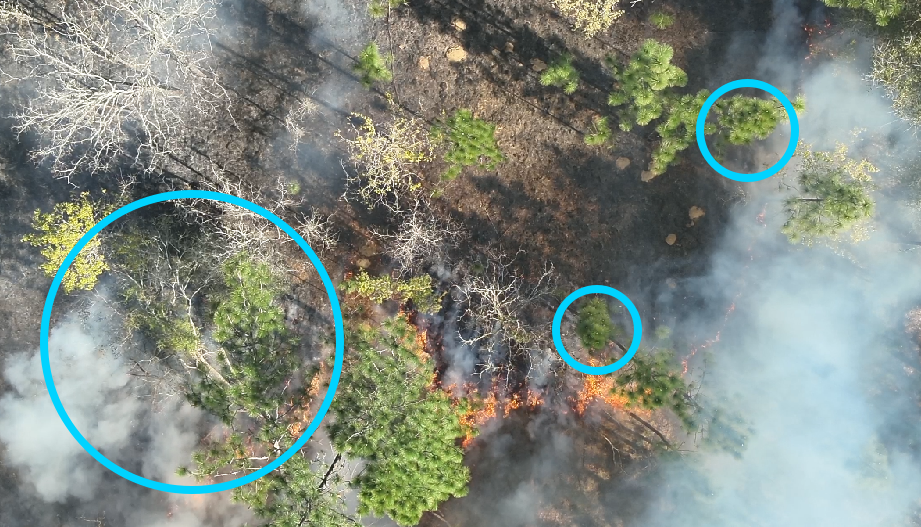}
\includegraphics[width=0.4\linewidth]{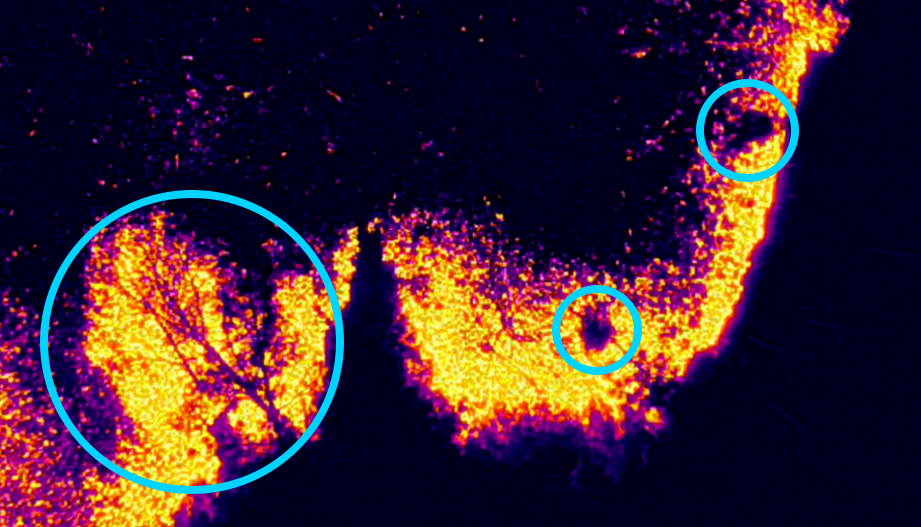}
\includegraphics[width=0.4\linewidth]{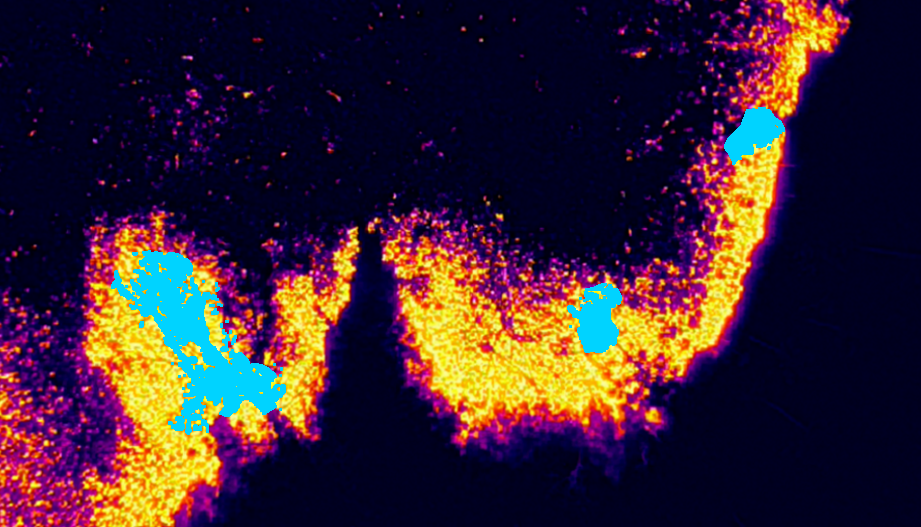}
\includegraphics[width=0.4\linewidth]{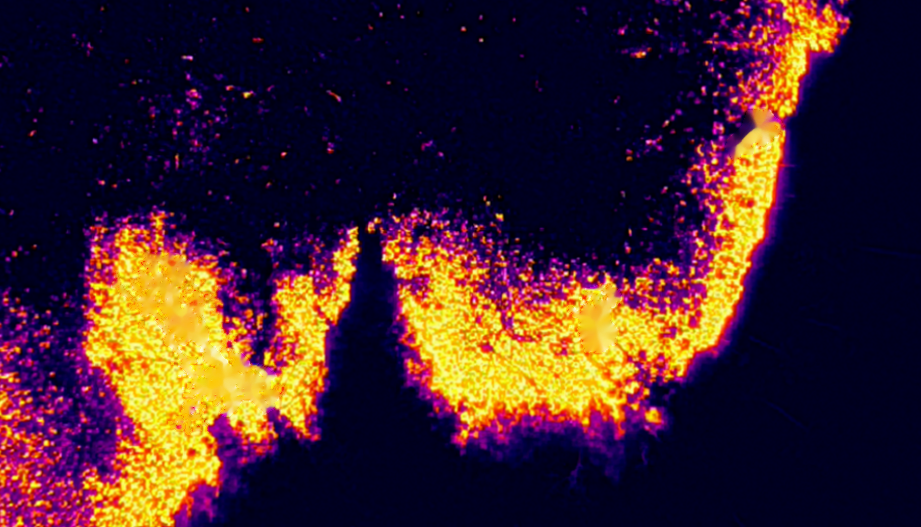}
\end{center}
  \caption{{\em Top:} A visual ({\em left}) and infrared ({\em right})
  image of a fire spreading from the top left corner to the bottom edge,
  captured via UAV. Both images are overlaid with blue circles to
  highlight the locations of notable occlusions from the infrared image.
  These circles are provided for clarity but do not play a role in the
  computations. {\em Bottom left:} The occlusion mask (denoted in blue)
  overlaid on the infrared image, found by identifying non-combusting
  pixels within the burning region using FDV. {\em Bottom right:} The
  inpainted infrared image.}
\label{fig:inpaint}
\end{figure*}

Figure~\ref{fig:inpaint} illustrates a sample application of image
inpainting to an infrared image of a fire obtained via UAV. In this
scenario, fire spreads from the top left corner of the image toward the
bottom edge. The UAV captures visual and infrared images side-by-side,
and a sample frame of each is shown in the top row of
Figure~\ref{fig:inpaint}. For easy reference and comparison, the sample
images are overlaid with blue circles that identify occlusions to be
inpainted in the infrared image. We see that the leftmost circle
contains a clear pattern of tree branches projected onto the surface
temperature, and can reasonably assume the infrared camera is measuring
the tree branch temperatures rather than the surface fuel temperatures.

As seen in Figure~\ref{fig:inpaint}, not all ``occlusions'' in the
visual image, such as small tree matter and smoke, correspond to missing
information in the infrared image. Visual and infrared imagery are
fundamentally different in what they capture, so a visual occlusion is
not always a strict indicator of an infrared one. As such, the most
appropriate way to identify occlusions depends on video features and
application. The occlusions in Figure~\ref{fig:inpaint} are identified
using FDV's segmentation step to locate non-combusting (pink, purple,
and dark blue) pixels within the flaming region, and this
straightforward approach is effective for the given example. While many
well-established methods for detecting trees and other objects in aerial
imagery exist~\cite{wang2018review, wulder2004comparison} and could be
useful for processing large datasets, selecting the right method is
scenario-dependent and beyond the scope of this work. For research
models that require extracting specific tree canopy features instead of
obscuring them, we recommend supplementing FDV's dataset generation with
machine learning approaches like those developed by
DeCastro~\etal~\cite{decastro2022computationally} and
Moran~\etal~\cite{moran2020mapping}. However, note that these approaches
are trained on satellite images of areas prior to fire and are not
specifically designed for fire videos.

%% ---------------------------
%% Pipeline
%% ---------------------------
\subsection{FDV Pipeline}
\label{sec:pipe}

The contents of the dataset FDV generates is determined by user input.
The exported dataset contains an array for each relevant set of
calculations performed. For example, an array with burning cell
information contains 0's for undisturbed cells and 1's for burning cells
at each timestep. If the desired data include burning cells, burned and
cooling cells, and smoke cell locations, the output array contains the
numbers 0 (undisturbed), 1 (burning), 2 (burned and cooling), and 3
(smoke). A summary of the FDV framework is provided in
Algorithm~\ref{alg:fdv}.

\begin{algorithm}
  \caption{FDV Algorithm}\label{alg:fdv}
  \begin{algorithmic}[1]
    \State Read video/image sequence
    \State User sets sampling frequency and parameter values for
    hierarchical cleaning (number of layers and radius-threshold value
    pairs), DBSCAN, and $\alpha$ radius
    \If{Visual image}
      \State Open test image and user selects a region of interest
      \State Segment selected region, convert image from RGB to HSV, and
      user reselects region of interest
    \ElsIf{Infrared image}
      \State Open test image and user selects a region of interest
    \EndIf
    \If{Inpainting}
      \State Reopen test image and user selects region for inpainting
    \EndIf
    \For {Every frame}
      \If{Inpainting}
	\State Segment and fill pixels identified for inpainting
      \EndIf
      \If{Visual image}
	\State RGB and HSV segmentation (thresholding)
      \ElsIf{Infrared image}
        \State Temperature segmentation (thresholding)
      \EndIf
      \State Hierarchical cleaning
      \If{Multiple region identification}
        \State Separate remaining points into distinct regions with
        DBSCAN
      \EndIf
	\State Return points connected to target regions
        \State Calculate $\alpha$-shape and extract boundary of each
        target region
      \EndFor
      \If{Velocity calculations}
        \For{Every frame}
	  \State Match boundary points between successive frames
	  \State Convert displacements to velocities
        \EndFor
      \EndIf
      \If{Statistical distribution analysis}
        \State Fit velocity distributions with MCMC
        \State Report fit parameters, error bounds, and normalized root
        mean square error (NRMSE)
      \EndIf
      \If{Dataset generation}
        \State Generate, format, and export dataset
      \EndIf
  \end{algorithmic} 
\end{algorithm}

%% ---------------------------
%% Discussion
%% ---------------------------
\section{Discussion} %% {{{
\label{sec:experiments}

FDV functions proficiently for visual and infrared videos---both
commonly used in fire science---allowing researchers to compare trends
between visual and infrared data for a fuller picture of the dynamics at
play. In this section, we showcase results for different combinations of
camera position, camera type, and measurement scale. We begin by showing
that MCMC outperforms moment matching when fitting statistical
distributions (Section~\ref{sec:MM}). We then consider an overhead view
of fire and plume spread that captures structures and movement in the
horizontal ($x$-$y$) plane. In this, we explore sub-microscale visual
and infrared images recorded adjacent to one another
(Section~\ref{sec:small_xy}). After, we analyze microscale visual
imagery capturing the vertical ($x$-$z$) plane (Section~\ref{sec:xz}).
To provide context for the reliabilty of FDV's calculations, we compare
our results to the output from other popular segmentation and tracking
methods (Section~\ref{sec:compare}). Finally, in
Section~\ref{sec:future}, we briefly review applications that have been
tested but not thoroughly analyzed, along with potential future
directions.

%%%%%%%%%%%%%%%%%%%%%%%%%%%%%%%%%%%%%%%%%%%%%%%%%%%%%%%%%%%%%%%%%%%%%%%
\subsection{Distribution Fitting}
\label{sec:MM}
\begin{figure*}[h]
\begin{center}
\includegraphics[width=0.45\linewidth]{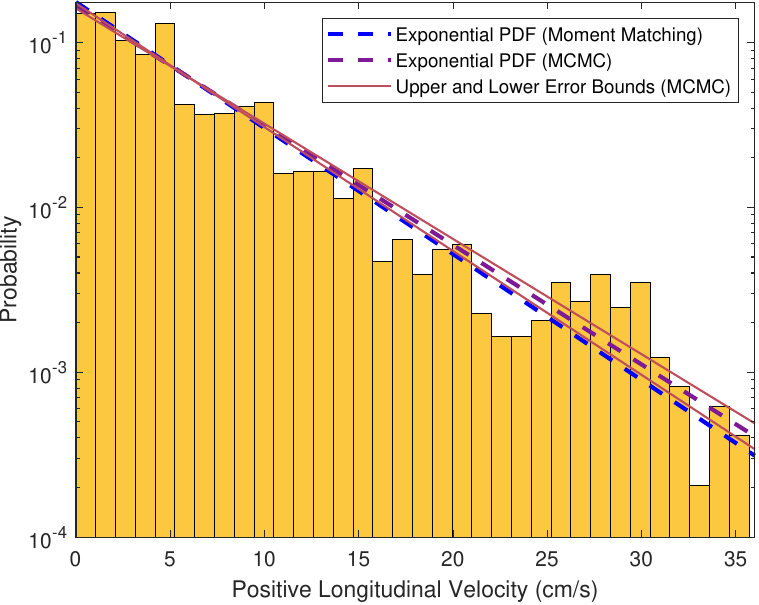}
\quad
\includegraphics[width=0.45\linewidth]{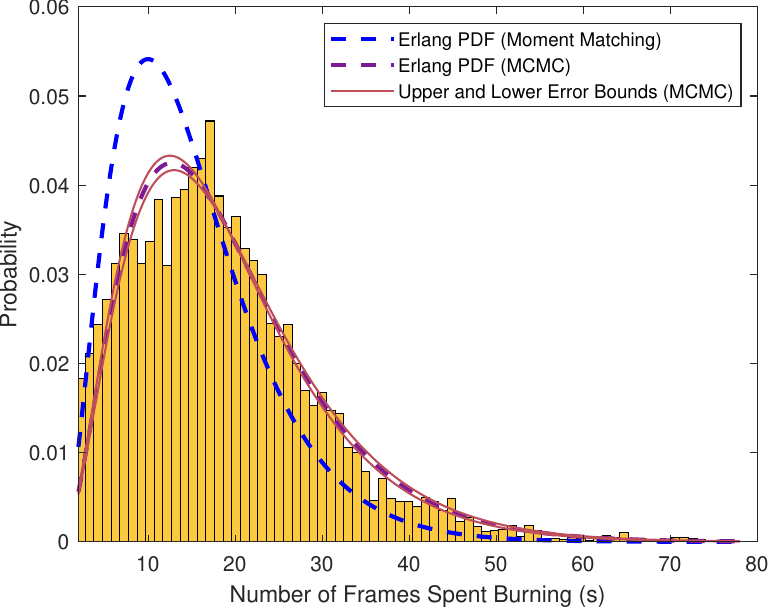}
\end{center}
   \caption{{\em Left:} Positive longitudinal ($L^+$) forward rate of
   spread distribution in semi-log scale with mean $\mu_{L^+}=5.68$~cm/s
   and standard deviation $\sigma_{L^+}=5.63$~cm/s. {\em Right:} Burn
   time distribution with mean $\mu = 18.5$~s and standard deviation
   $\sigma = 11.1$~s. The blue curves are found with moment matching and
   the purple curves are found with MCMC. The normalized root mean
   square error between the curves and the data is reported in
   Table~\ref{tab:mcmc}. Plots reproduced and modified with
   permission~\cite{sagel2021fine}.}
\label{fig:oldIRdists}
\end{figure*}

We apply MCMC and moment matching to two statistical distributions from
our previous work: the positive longitudinal velocity and the burn
time~\cite{sagel2021fine}. Figure~\ref{fig:oldIRdists} shows the
distributions. The blue line is found with moment matching, while the
purple lines use MCMC. The uncertainty of the MCMC distribution is
illustrated with the solid lines. The burn time distribution clearly shows that MCMC outperforms
moment matching, in addition to providing error bounds. Therefore, we use MCMC for the remainder of the paper.

Table~\ref{tab:mcmc} summarizes the normalized root mean square error
(NRMSE) associated with velocity (exponential) and burn time (Erlang)
distributions. Fitting these distributions with MCMC results in a NRMSE
that is up to a factor of two times lower than the NRMSE associated with
values obtained through moment matching. This expanded statistical
analysis is the start of a new framework within FDV that finds the best
continuous distribution fit for given data. Physical interpretations of
fire behavior and the application of vision-based research to fire and
plume models are both strongly influenced by the type of distribution
determined as a best fit to the data.

\begin{table}[ht]
  \begin{center}
    {\small{
\begin{tabular}{llll}
\toprule
Source & Method & Distribution & NRMSE \\
\midrule
Infrared & Moment-Matching & Exponential & $2.31 \times 10^{-4}$ cm/s\\
Infrared & MCMC & Exponential & $2.23 \times 10^{-4}$ cm/s\\
Infrared & Moment-Matching & Erlang & $9.03 \times 10^{-5}$ s\\
Infrared & MCMC & Erlang & $4.23 \times 10^{-5}$ s\\
\bottomrule
\end{tabular}
}}
\end{center}
\caption{A comparison of the normalized root mean square error (NRMSE)
  associated with each distribution fitting method. The MCMC approach
  results in an error that is twice as small as moment
  matching.}
\label{tab:mcmc}
\end{table}

%%%%%%%%%%%%%%%%%%%%%%%%%%%%%%%%%%%%%%%%%%%%%%%%%%%%%%%%%%%%%%%%%%%%%%%
\subsection{Sub-Microscale $x$-$y$ Fire Spread}
\label{sec:small_xy}

\begin{figure*}[h]
\begin{center}
\includegraphics[width=0.9\linewidth]{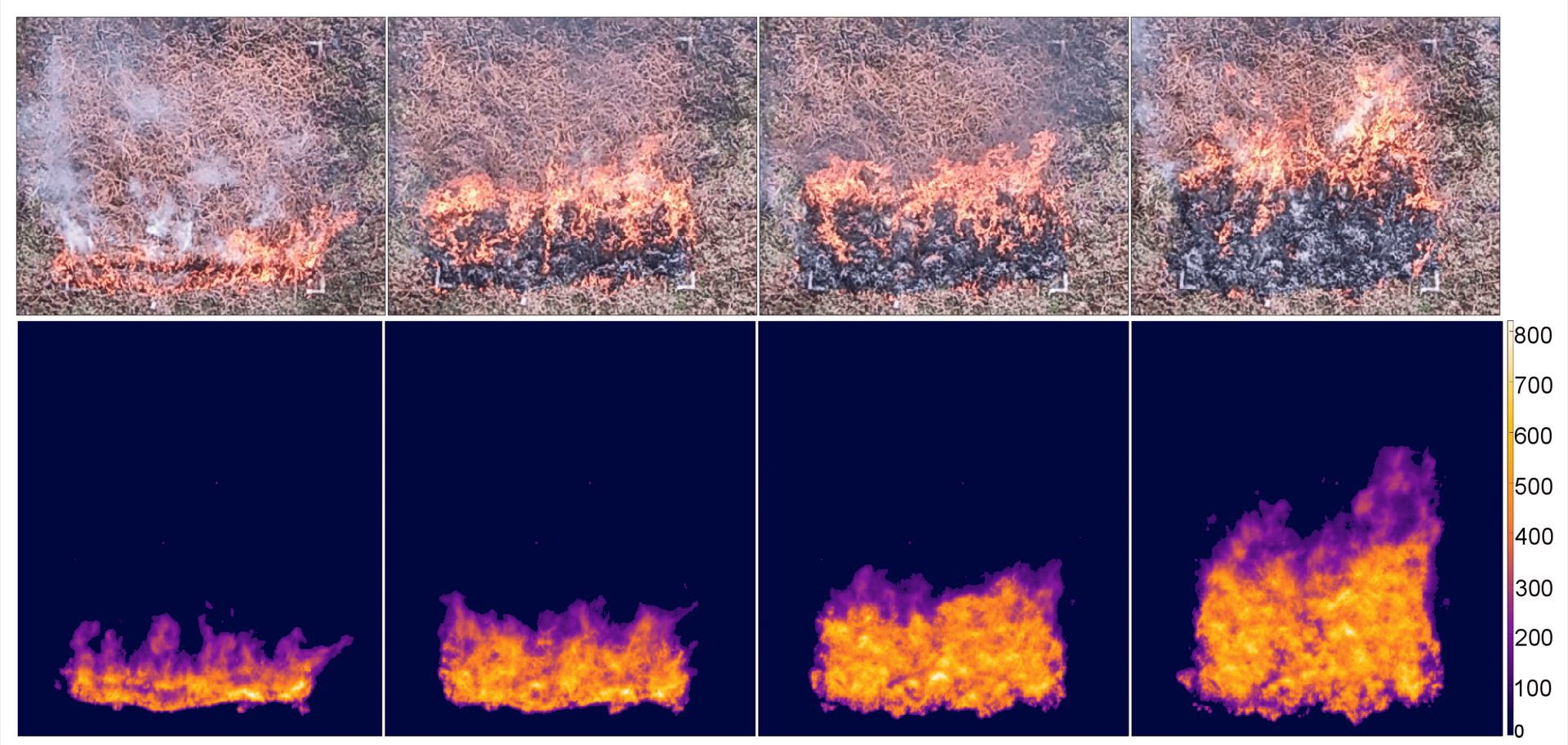}
\end{center}
   \caption{Visual ({\em top}) and infrared ({\em bottom}) camera frames at four
   timesteps spaced approximately 10~s apart. The frames of the
   individual cameras are not perfectly synchronized. Figure reprinted
   with permission~\cite{sagel2021fine}.}
\label{fig:oldfig}
\end{figure*}

This experiment builds on our previous work~\cite{sagel2021fine}, where
we analyzed fire spread across a 2~m~$\times$~2~m quasi-homogeneous pine
straw fuel bed using infrared data. Although we captured visual images
of the fire with a Hero 7 GoPro mounted alongside the Teledyne FLIR
A655sc infrared camera, our original algorithm could not effectively
process visual imagery of fires or plumes. With the updated FDV
algorithm, we now process both the visual and infrared components of
this dataset and compare the calculated trends with those found using
the old methodology. For reference, sample visual and infrared frames
are shown in Figure~\ref{fig:oldfig}.

%%%%%%%%%%%%%%%%%%%%%%%%%%%%%%%%%%%%%%%%%%%%%%%%%%%%%%%%%%%%%%%%%%%%%%%
\subsubsection{FDV Applied to Infrared Data}

\begin{figure*}[htp!]
\begin{center}
\includegraphics[width=0.22\linewidth]{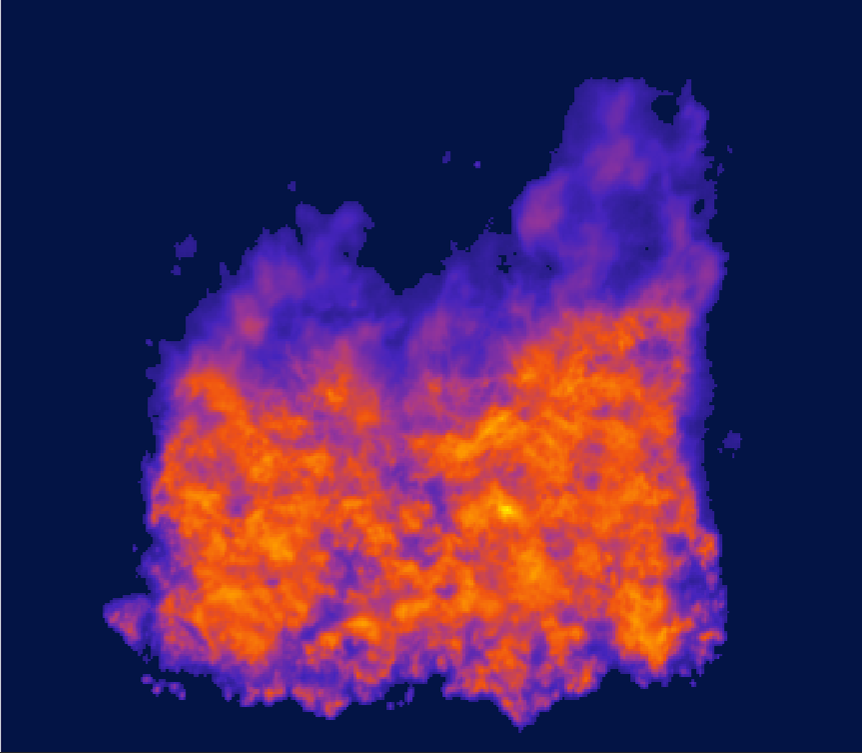}
\includegraphics[width=0.22\linewidth,trim=0cm 0.1cm 0cm 0cm,clip=true]{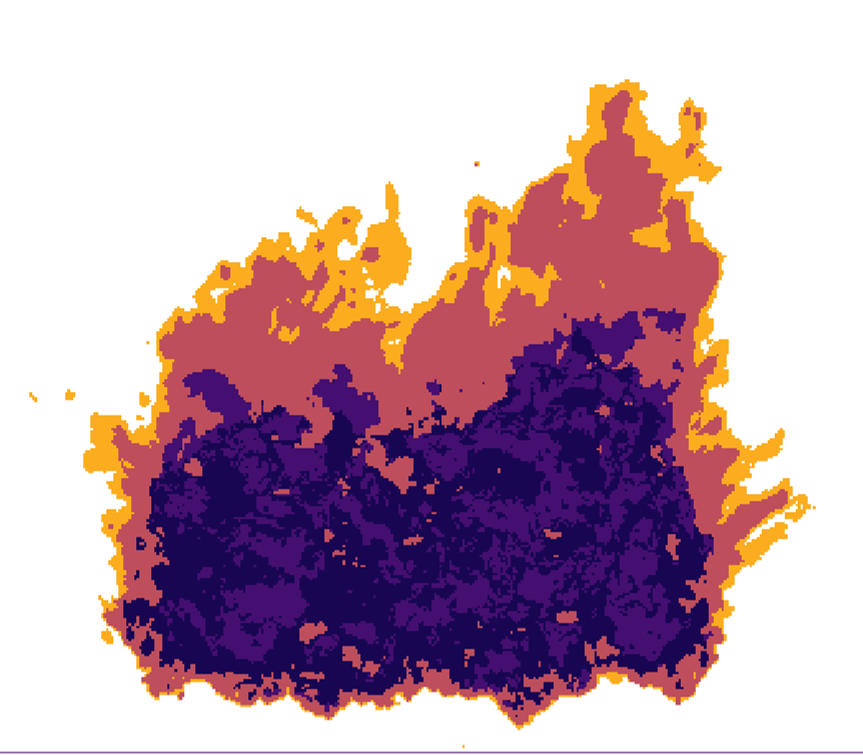}
\includegraphics[width=0.22\linewidth]{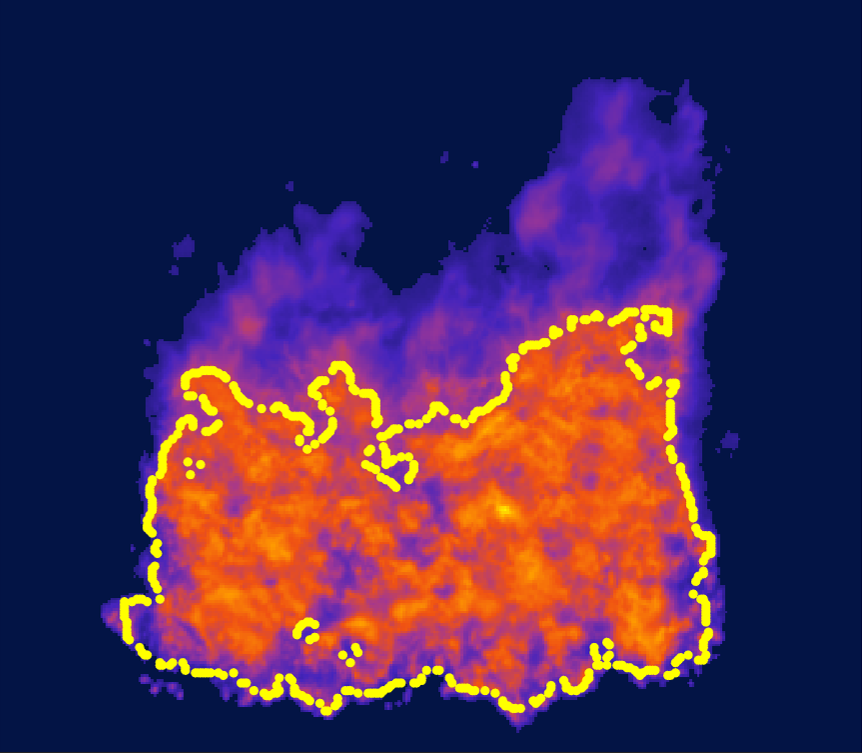}
\includegraphics[width=0.22\linewidth]{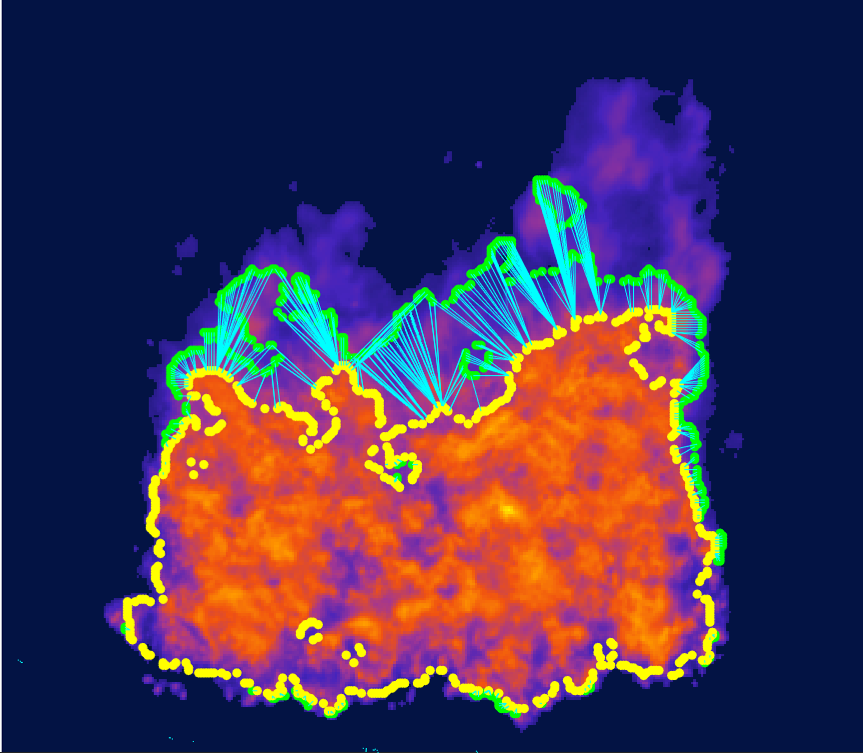}
\end{center}
\caption{Visualization of a segment of the FDV pipeline applied to
  infrared images. From left to right, the figure shows: the original
  infrared image, the temperature-segmented image, the boundary
  of the burned and cooling region (in contrast to the flaming region),
  and the displacement between this boundary and the boundary in the next timestep.}
\label{fig:xyvisualization}
\end{figure*}

\begin{figure*}[htp!]
\begin{center}
\includegraphics[width=0.3\linewidth]{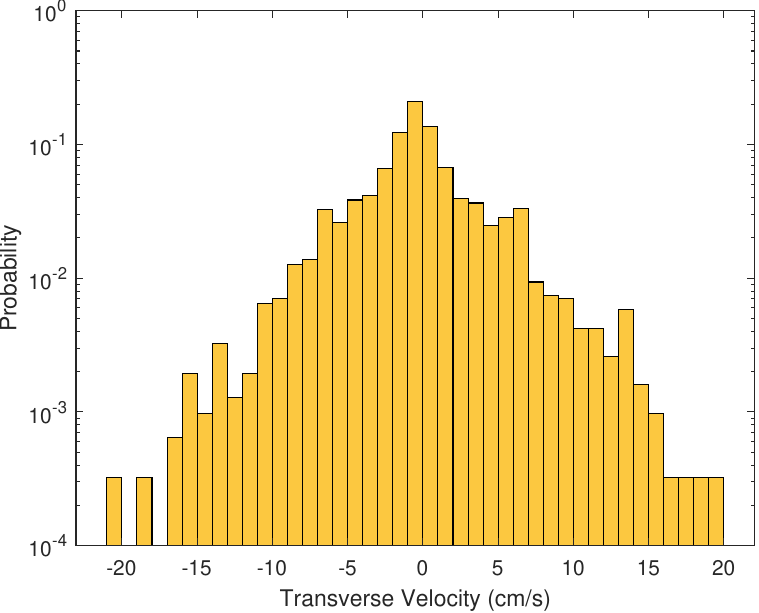}
\quad
\includegraphics[width=0.3\linewidth]{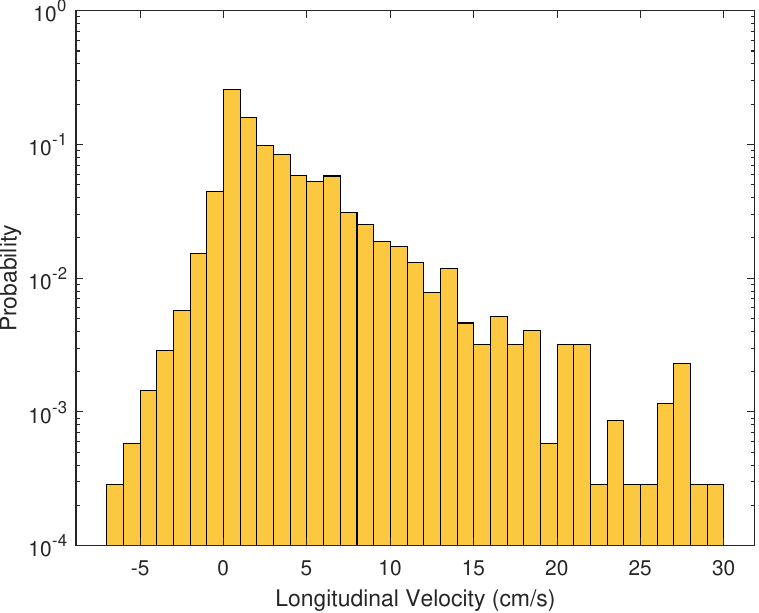}
\quad
\includegraphics[width=0.3\linewidth]{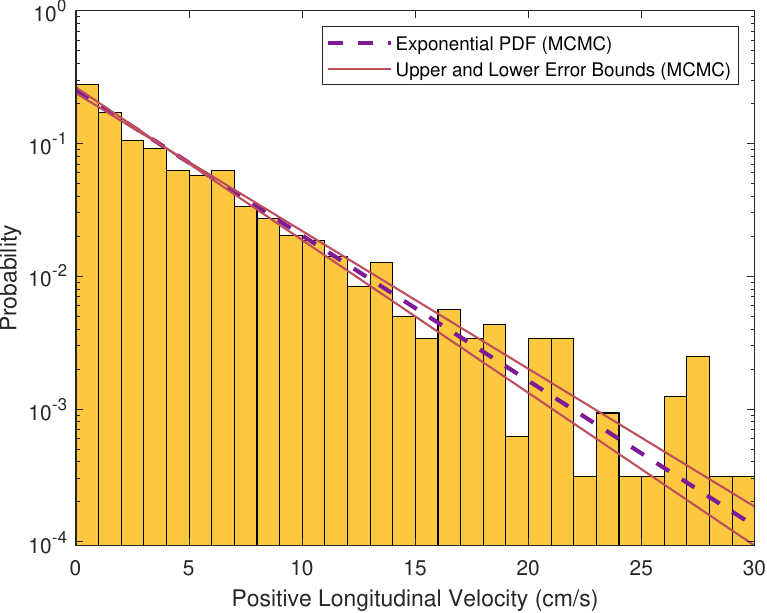}
\end{center}
\caption{\label{fig:newIRdists} Velocity distributions for results
  obtained processing the same infrared dataset as in our previous
  work~\cite{sagel2021fine} using FDV. {\em Left:} Transverse velocity
  distribution. {\em Middle:} Longitudinal velocity distribution. {\em
  Right:} Positive longitudinal forward rate of spread distribution
  along with the best fit exponential curve. The means, standard
  deviations, best fit line, and NRMSE are in the second row of
  Table~\ref{tab:infraredStats}.}
\end{figure*}

We reprocess the same set of infrared images with the updated FDV
pipeline; a visualization of the updated processing applied to a sample
infrared image is shown in Figure~\ref{fig:xyvisualization}. A complete
set of visualized results for all infrared frames is provided in the
\href{https://github.com/quaife/FDV_Data}{Supplementary Material}.
Parsing the results frame-wise provides visual verification---a common
metric for assessing the accuracy of computer vision output without a
labeled dataset---that FDV performs reliably in this application. The
statistical results determined used all infrared frames are displayed in
Figure~\ref{fig:newIRdists}. Our choice to use greedy nearest neighbor
matching, rather than the modified assignment problem, causes the
updated distributions to feature slightly lower maximum positive
longitudinal velocity values. The updated distributions feature slightly
lower maximum positive longitudinal velocity values, likely caused by
our implementation of greedy nearest neighbor matching rather than use
of the modified assignment problem. The means and standard deviations of
the longitudinal, transverse, and positive longitudinal rates of spread
are reported in Table~\ref{tab:infraredStats}. Overall, the
distributions have similar shape and statistics. Though the NRMSE is
twice as large for the new FDV methodology, they are both on the scale
of $10^{-4}$~cm/s; this is a small difference relative to the scale of
the actual motions. Additionally, calculating the root mean square error
(RMSE) between the binned velocity data and exponential distribution fit
for each individual bin reveals that the FDV results produce a lower
error than the old methodology in two-thirds of the bins.

\begin{table}[htp]
  \centering
  \begin{tabular}{c|cccccccc}
    & $\mu_{L}$ & $\sigma_L$ & $\mu_T$ & $\sigma_T$ & $\mu_{L^+}$ &
    $\sigma_{L^+}$ & $\lambda$ & NRMSE \\
    \hline
    Old & 4.60 & 6.22 & 0.74 & 7.43 & 5.68 & 5.63 & 
      $1.76 \times 10^{-1}$ & $2.31 \times 10^{-4}$ \\
    FDV & 3.70 & 4.30 & -0.32 & 4.30 & 4.11 & 4.24 & 
      $(2.52 \pm 0.13) \times 10^{-1}$ & $4.55 \times 10^{-4}$ 
  \end{tabular}
  \caption{\label{tab:infraredStats} The means ($\mu$) and standard
  deviations ($\sigma$) of the longitudinal velocity ($L$), transverse
  velocity ($T$), and positive longitudinal velocity ($L^+$), all with
  units cm/s. Also reported are the exponential fit parameters,
  $\lambda$, in s/cm and the associated NRMSE in cm/s. The data in the
  first row use our old methodology~\cite{sagel2021fine}, and the data
  in the second row use the techniques described in this paper. Unlike
  our old method, FDV includes error bounds for $\lambda$.}
\end{table}

%%%%%%%%%%%%%%%%%%%%%%%%%%%%%%%%%%%%%%%%%%%%%%%%%%%%%%%%%%%%%%%%%%%%%%%
\subsubsection{Comparing Infrared and Visual Data}
\label{sec:IRandVIS}
Since the infrared video is captured at 1 Hz, we apply FDV to the visual
video sampled at 1 Hz and compare the two distributions. Note, however,
that the infrared video captures isotherms, whereas the visual video
captures flame location. These are fundamentally different values,
though we can look at both as a representation of the fire front's
forward propagation. Additionally, with the infrared data, we are
isolating and tracking a single boundary between the burned and burning
regions (i.e., one curve spanning the plot), but for visual we have both
the upper and lower boundaries of the flaming region, leading to a
greater number of points and associated velocity measurements. 

\begin{figure*}[h]
\begin{center}
\includegraphics[width=0.3\linewidth]{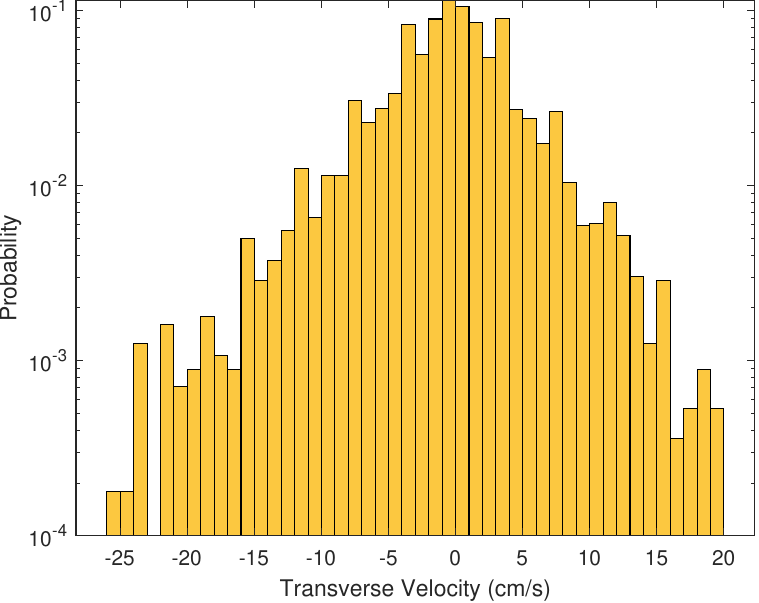}
\quad
\includegraphics[width=0.3\linewidth]{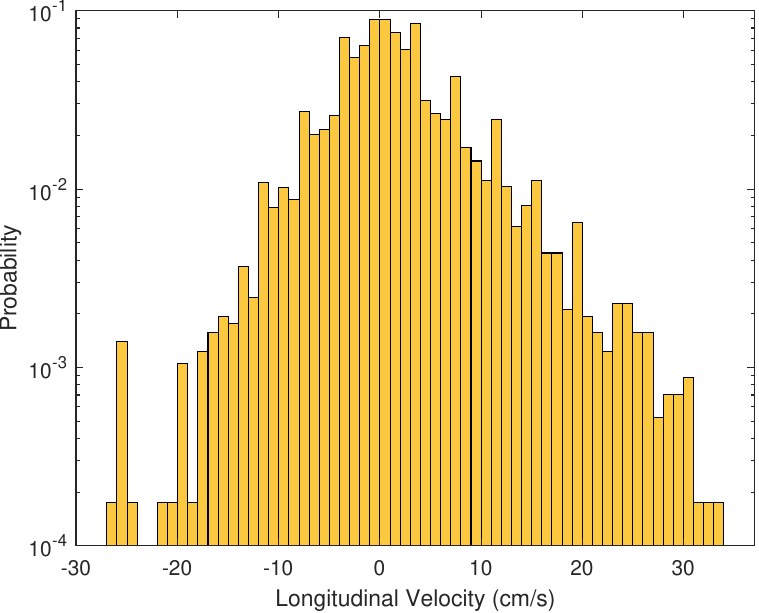}
\quad
\includegraphics[width=0.3\linewidth]{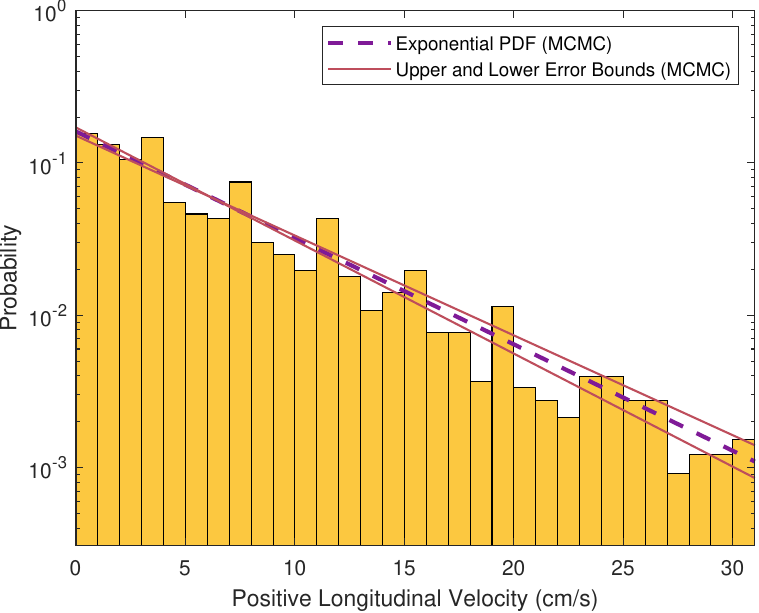} \\
\includegraphics[width=0.3\linewidth]{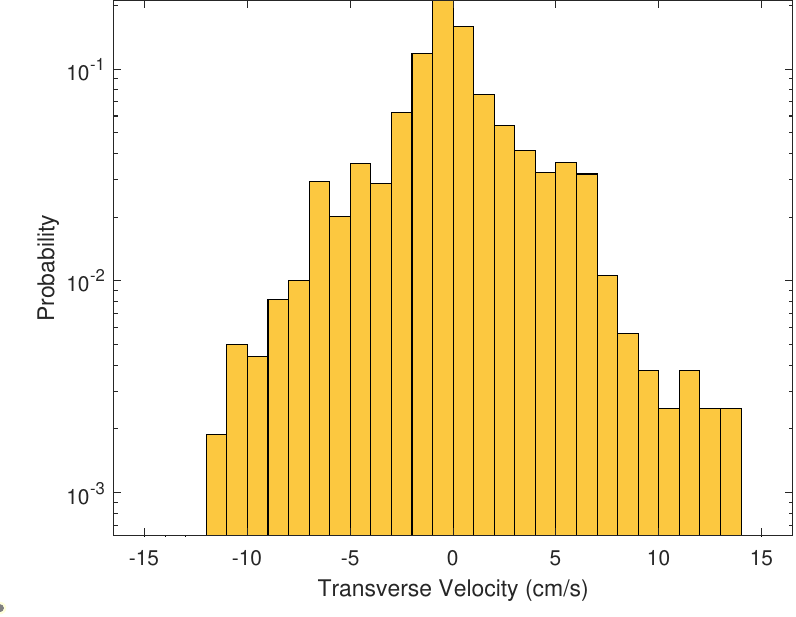}
\quad
\includegraphics[width=0.3\linewidth]{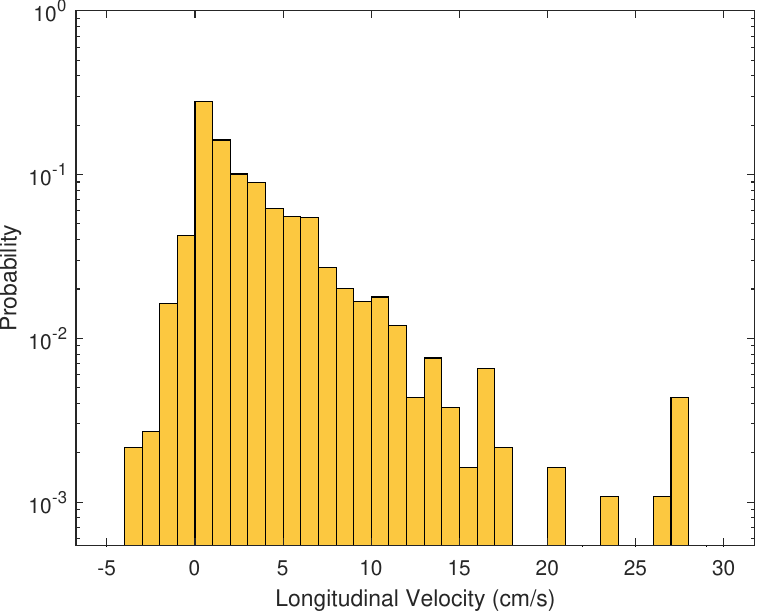}
\quad
\includegraphics[width=0.3\linewidth]{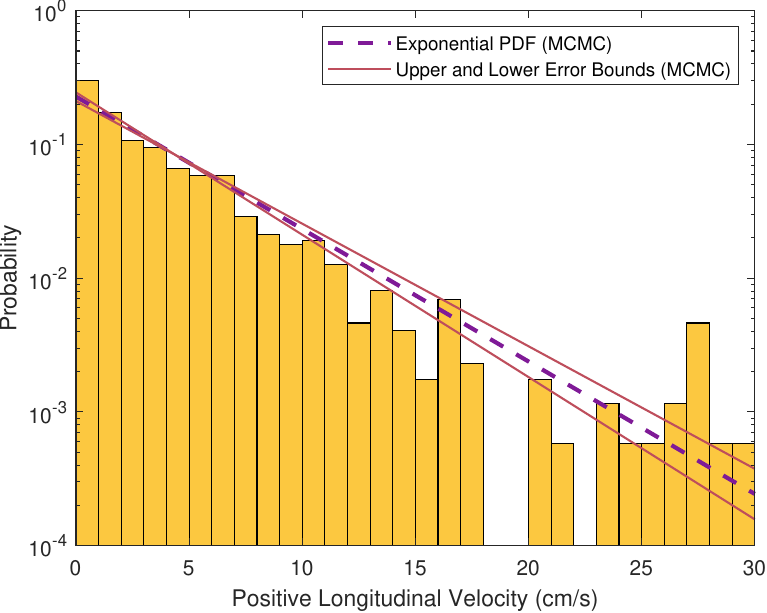}
\end{center}
   \caption{\label{fig:newVISdists1hz_timecrop} Velocity distributions
   for results obtained processing the visual image dataset sampled at
   1~Hz ({\em top}) and results obtained subsampling the infrared image
   dataset at the same timesteps as the visual 1~Hz data ({\em
   bottom}). The means, standard deviations, best fit lines, and NRMSE
   values are in Table~\ref{tab:VisIRStats}.}
\end{figure*}

The velocity results for the visual video are shown in the top row of
Figure~\ref{fig:newVISdists1hz_timecrop}. Contrary to the infrared
video, the visual flames are contained entirely within the 2~m $\times$
2~m plot for only 24~s after the ignition pattern resolves. Beyond this
time, the flames frequently extend beyond the field of view. To make a
fair comparison between the infrared and visual distributions, we
crop the infrared data at the same 24~s as the visual distribution.
This cropped distribution is in the bottom row of
Figure~\ref{fig:newVISdists1hz_timecrop}, and we note that this
distribution is simply a subsample of the distribution in
Figure~\ref{fig:oldIRdists}. 

\begin{table}[htp]
  \centering
  \begin{tabular}{c|cccccccc}
    & $\mu_{L}$ & $\sigma_L$ & $\mu_T$ & $\sigma_T$ & $\mu_{L^+}$ &
    $\sigma_{L^+}$ & $\lambda$ & NRMSE \\
    \hline
    Visual & 1.54 & 6.88 & -0.43 & 5.32 & 5.78 & 5.50 & 
      $(1.61 \pm 0.10) \times 10^{-1}$ & $3.87 \times 10^{-4}$ \\
    Infrared & 3.45 & 4.08 & $7.60 \times 10^{-3}$ & 3.71 & 
      3.77 & 4.02 & $(2.28 \pm 0.17) \times 10^{-1}$ & $1.01 \times 10^{-3}$ 
  \end{tabular}
  \caption{\label{tab:VisIRStats} A comparison of the distributions
  computed with FDV when applied to visual 1~Hz data and infrared 1~Hz
  data, both at the same 24~s. The means ($\mu$) and standard deviations
  ($\sigma$) of the longitudinal velocity ($L$), transverse velocity
  ($T$), and positive longitudinal velocity ($L^+$), all with units
  cm/s. Also reported are the exponential fit parameters, $\lambda$, in
  s/cm and the associated NRMSE in cm/s.}
\end{table}

When comparing the two distributions, it is apparent that the visual
results contain larger positive velocities and a greater number of
negative velocities than their infrared counterparts. This is
understandable, as visual video captures the flames' oscillation, unlike
infrared video which only indicates the flames' oscillatory movement
when a flame has laid down long enough to sufficiently heat the fuel
around it. This laying down behavior has been researched and established
in existing literature~\cite{gorham2014studying} and is consistent with
trends observed in our previous analysis of the infrared
images~\cite{sagel2021fine}. Additionally, in the visual video, we have
the effect of variable flame visibility despite the gas in that area
still being heated. This is colloquially referred to as the flames'
flickering. Flickering leads to sudden jumps in the position of the
flame---regardless of the sampling frequency used to capture the fire's
motion---that are not captured in the infrared imagery. However, we see
from the distributions (Figure~\ref{fig:newVISdists1hz_timecrop}) and
associated statistics (Table~\ref{tab:VisIRStats}) that all of these
newly calculated motions from the visual video occur on relatively the
same spatial scale and order of magnitude as motions calculated from the
infrared imagery.

Effects of the aforementioned behaviors are also present in the
statistics provided in Table~\ref{tab:VisIRStats}. The existence of
larger positive velocities in the visual data is evident in the mean
positive longitudinal velocity, $\mu_{L^+}$, where the visual mean value
is greater than the infrared mean value. Conversely, the larger negative
velocities are attributable primarily to flame flickering. Since the
visual flames are often seen laying down due to wind and their natural
oscillations, the sudden disappearance of visual flame inevitably causes
the displacement between boundaries at successive timesteps to contain a
large backwards motion toward the remaining body of flames.

%%%%%%%%%%%%%%%%%%%%%%%%%%%%%%%%%%%%%%%%%%%%%%%%%%%%%%%%%%%%%%%%%%%%%%%
\subsubsection{Comparing Visual Data at Different Sampling Rates}
\label{sec:samplingrate} Unlike the infrared video, which is recorded at
1 Hz, the visual video is recorded at 30 Hz. This allows us to subsample
the frames to capture dynamics at different scales. Sampling at too
low a frequency will miss some structures and dynamics, especially
because we are measuring the fire's behavior at a fine spatial scale.  Moreover, sampling at too high a
frequency will amplify the inevitable noise in the data and turbulence
in the flame dynamics. We can quickly rule out sampling at some of the
largest available frequencies by considering the dimensions of our
experiment. The video has a spatial resolution of $0.79$~cm/px, so a
displacement of one pixel measured at $30$~Hz results in a minimum
measurable velocity of $23.7$~cm/s. Such a large minimum velocity is
unphysical since it would result in the flames traveling nearly three
times the length of the entire burn unit in the 24~s experiment. This
same argument also rules out other large sampling frequencies. In
particular, any sampling rate larger than $10$~Hz results in unphysical
velocities.

We apply the Nyquist-Shannon sampling theorem to further eliminate
remaining sampling rates. In particular, we use the fact that the
sampling frequency, $f$, should be two times larger than the bandwidth
of the signal so that aliasing is avoided. Given the image resolution
($\mathtt{RES}$) and field of view ($\mathtt{FOV}$), the 
maximum measurable velocity ($u_{\max}$) is
\begin{align*}
  u_{\max} = \frac{f}{2} \times \frac{\mathtt{FOV}}{\mathtt{RES}}.
\end{align*}
The image resolution is $\mathtt{RES} = 1.27$~px/cm and the field of
view is $\mathtt{FOV} = 253$~px, which is the number of pixels along the
primary direction of spread. These values simplify the relationship
between sampling frequency and maximum measurable speed to 
\begin{equation*} 
  u_{\max} = 99.6f,
\end{equation*}
where $f$ is measured in Hz and $u_{\max}$ is measured in cm/s.

In Table~\ref{tab:shannon}, we report the maximum measurable velocity
($u_{\max}$) and maximum observed velocity ($u_{\mathrm{obs}}$) for
several sampling rates ($f$). To obtain $u_{\mathrm{obs}}$, FDV
processes the video at each sampling rate and the largest velocities are
recorded. We see that for all frequencies greater than 1--2~Hz, the
ratio between the maximum observed velocity and maximum measurable
velocity remains within a nearly constant range.  This indicates that at
these high frequencies, the flames are moving a fixed number of pixels
between frames independent of the frame rate.  Therefore, sampling
frequencies above 2~Hz artificially inflate the velocities, and we only
consider sampling rates of 1~Hz and 2~Hz for the reminder of this
analysis. Other experiments have found similar maximum flame
speeds~\cite{clements2007observing, finney2015role,
singh2023intermittent}, though the exact values vary slightly from ours
due to experimental setup.
\begin{table}[htp]
  \begin{center}
  {\small{
  \begin{tabular}{cccc}
    \toprule
$f$ (Hz)  & $u_{\max}$ (cm/s) & $u_{\mathrm{obs}}$ (cm/s) & Ratio \\
\midrule
1 & 99.6  & 33.3  & 33\% \\
2 & 199.2 & 90.5  & 45\% \\
5 & 498.0 & 226.2 & 45\% \\
\bottomrule
  \end{tabular}
  }}
  \end{center}
  \caption{\label{tab:shannon} The maximum measurable velocity
  ($u_{\max}$) and maximum observed velocity ($u_{\mathrm{obs}}$) at
  four sampling rates. Also reported is the ratio between the observed
  velocities and the maximum measurable velocities. Beyond 5~Hz, this
  ratio remains close to $45\%$.}
\end{table}

\begin{figure*}[htp!]
\begin{center}
\includegraphics[width=0.3\linewidth]{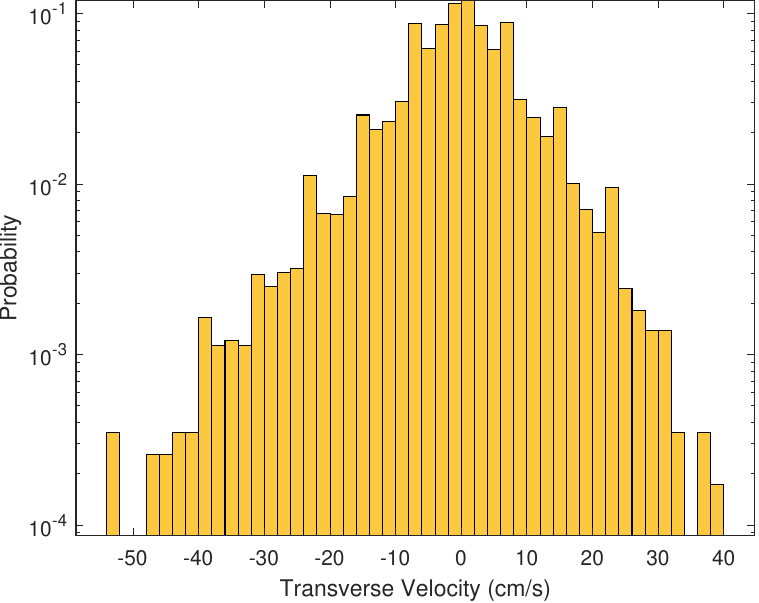}
\quad
\includegraphics[width=0.3\linewidth]{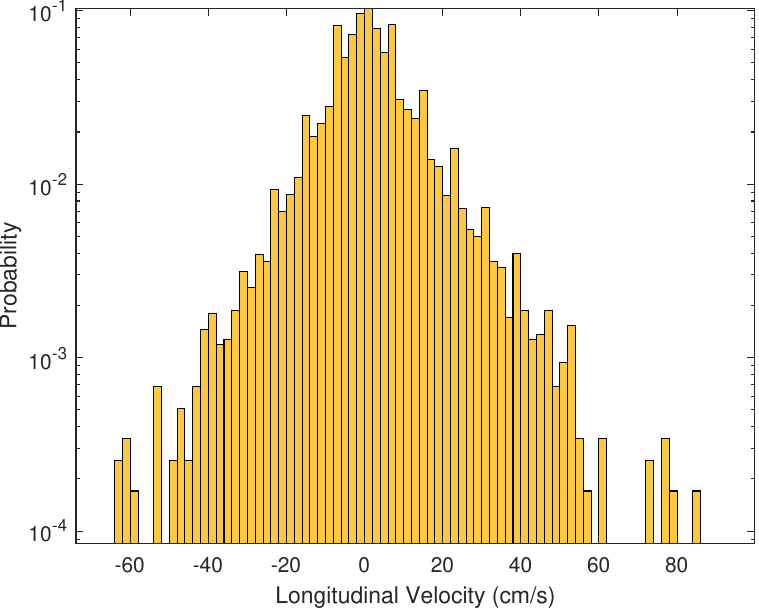}
\quad
\includegraphics[width=0.3\linewidth]{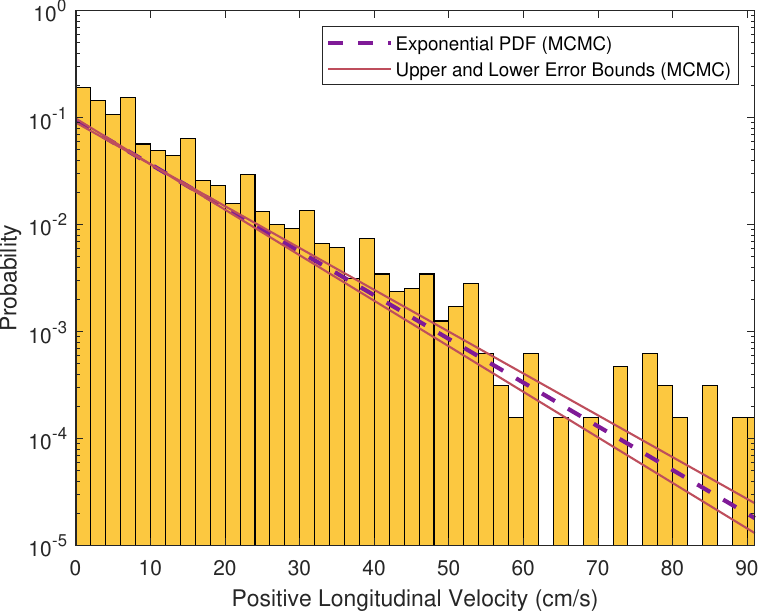}
\end{center}
  \caption{Velocity distributions for results obtained processing the
  visual image dataset sampled at 2~Hz using FDV. {\em Left:} Transverse
  velocity distribution in semi-log scale with mean $\mu=-0.35$~cm/s and
  standard deviation $\sigma=9.80$~cm/s. {\em Middle:} Longitudinal
  velocity distribution in semi-log scale with mean $\mu=1.60$~cm/s and
  standard deviation $\sigma=13.26$~cm/s. {\em Right:} Positive
  longitudinal forward rate of spread distribution in semi-log scale
  with mean $\mu=10.15$~cm/s and standard deviation $\sigma=10.36$~cm/s.
  An exponential curve is fit to this distribution with rate parameter
  $\lambda=0.094$ and NRMSE $1.24 \times 10^{-4}$~cm/s.}
  \label{fig:newVISdists2hz_timecrop}
\end{figure*}

The 2~Hz visual velocity distributions are provided in
Figure~\ref{fig:newVISdists2hz_timecrop}, with associated statistics in
the bottom row of Table~\ref{tab:VisualStats}. The 1~Hz visual velocity
statistics are repeated in the top row for easy comparison. Despite
having larger maximum and minimum values, the overall shape of the 2~Hz
distributions are very similar to the shape of the 1~Hz distributions in
Figure~\ref{fig:newVISdists1hz_timecrop}. The similarity in overall
shape is represented in the statistics, with nearly identical mean
values for both longitudinal velocities, $\mu_{L}$, and transverse
velocities, $\mu_T$. The maximum and minimum limits are nearly doubled
for the 2~Hz results, and this is also apparent in the statistics; the
standard deviations for longitudinal velocities, $\sigma_L$, and
transverse velocities, $\sigma_T$, are approximately doubled for the
2~Hz statistics compared to the 1~Hz statistics. We also see the effect
of this doubling in the positive longitudinal velocity, where the mean
and standard deviation are both doubled due to the exclusion of negative
values. The fitted distributions for the 1~Hz and 2~Hz sampling
frequencies have small NRMSE values, indicating that both datasets are
approximately exponentially distributed. This suggests that we are
observing similar underlying trends in fire spread independent of
sampling frequency. A complete set of visualized results for all visual
frames sampled at 2~Hz is provided in the
\href{https://github.com/quaife/FDV_Data}{Supplementary Material}.

\begin{table}[htp!]
  \centering
  \begin{tabular}{c|cccccccc}
    & $\mu_{L}$ & $\sigma_L$ & $\mu_T$ & $\sigma_T$ & $\mu_{L^+}$ &
    $\sigma_{L^+}$ & $\lambda$ & NRMSE \\
    \hline
    Visual (1~Hz) & 1.54 & 6.88 & -0.43 & 5.32 & 5.78 & 5.50 & 
      $(1.61 \pm 0.10) \times 10^{-1}$ & $3.87 \times 10^{-4}$ \\
    Visual (2~Hz) & 1.60 & 13.26 & -0.35 & 9.80 & 10.15 & 10.36 & 
      $(9.40 \pm 0.40) \times 10^{-2}$ & $1.24 \times 10^{-4}$ 
  \end{tabular}
  \caption{\label{tab:VisualStats} A comparison of the distributions
  computed with FDV when applied to visual data at 1~Hz and 2~Hz. The
  means ($\mu$) and standard deviations ($\sigma$) of the longitudinal
  velocity ($L$), transverse velocity ($T$), and positive longitudinal
  velocity ($L^+$), all with units cm/s. Also reported are the
  exponential fit parameters, $\lambda$, in s/cm and the associated
  NRMSE in cm/s.}
\end{table}

Though not included in our final analysis, we explored determining ideal
video sampling frequency using power spectra of the isotherm's position
along a single line of the plot. Visual interpretation of the results
suggested an ideal sampling frequency between $3$--$10$~Hz, but did so
without accounting for the two-dimensional structure or motions present.
Ultimately, we decided this approach made too many simplifications
to function as a tool for determining ideal sampling frequency. Analyzing
the results of more complex multidimensional data may achieve the
desired goal, but falls outside the scope of this work.

%%%%%%%%%%%%%%%
\subsection{Microscale $x$-$z$ Plume Evolution}
\label{sec:xz}

\begin{figure*}[h]
  \begin{center}
  \includegraphics[width=0.45\linewidth]{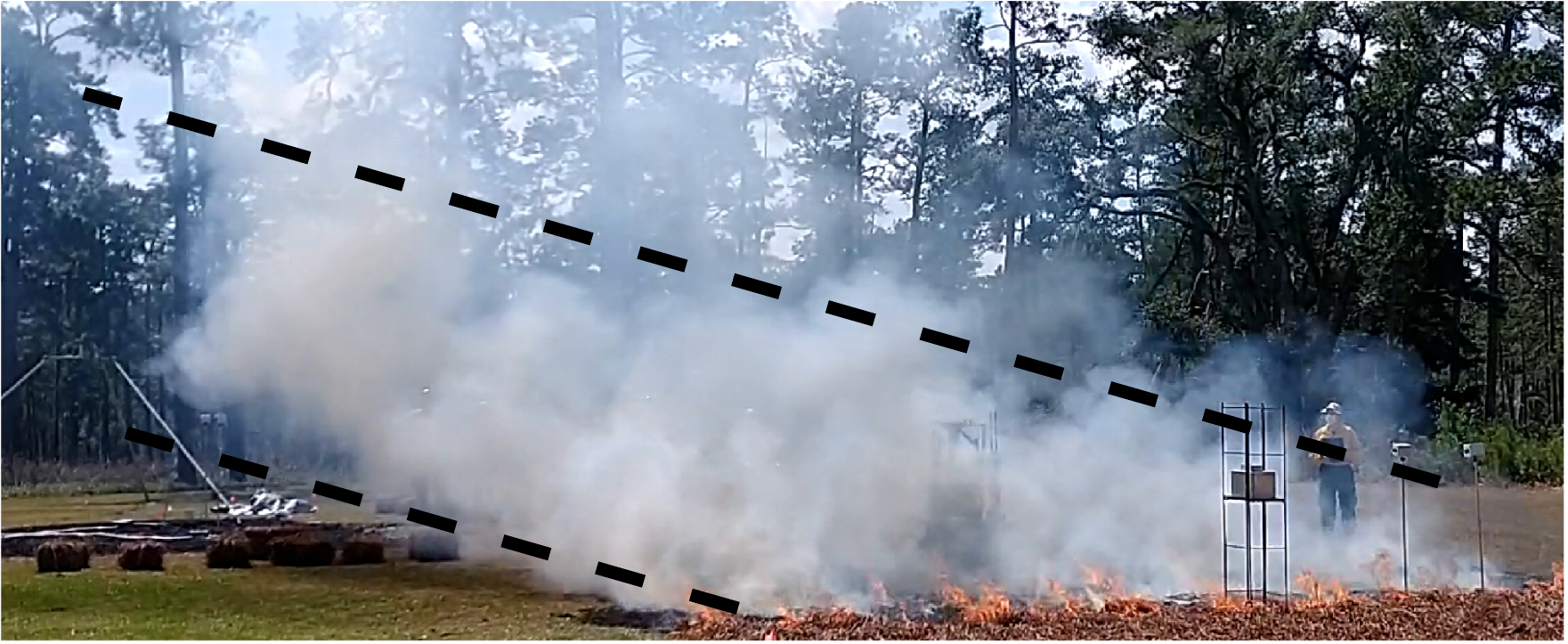}
  \includegraphics[width=0.45\linewidth]{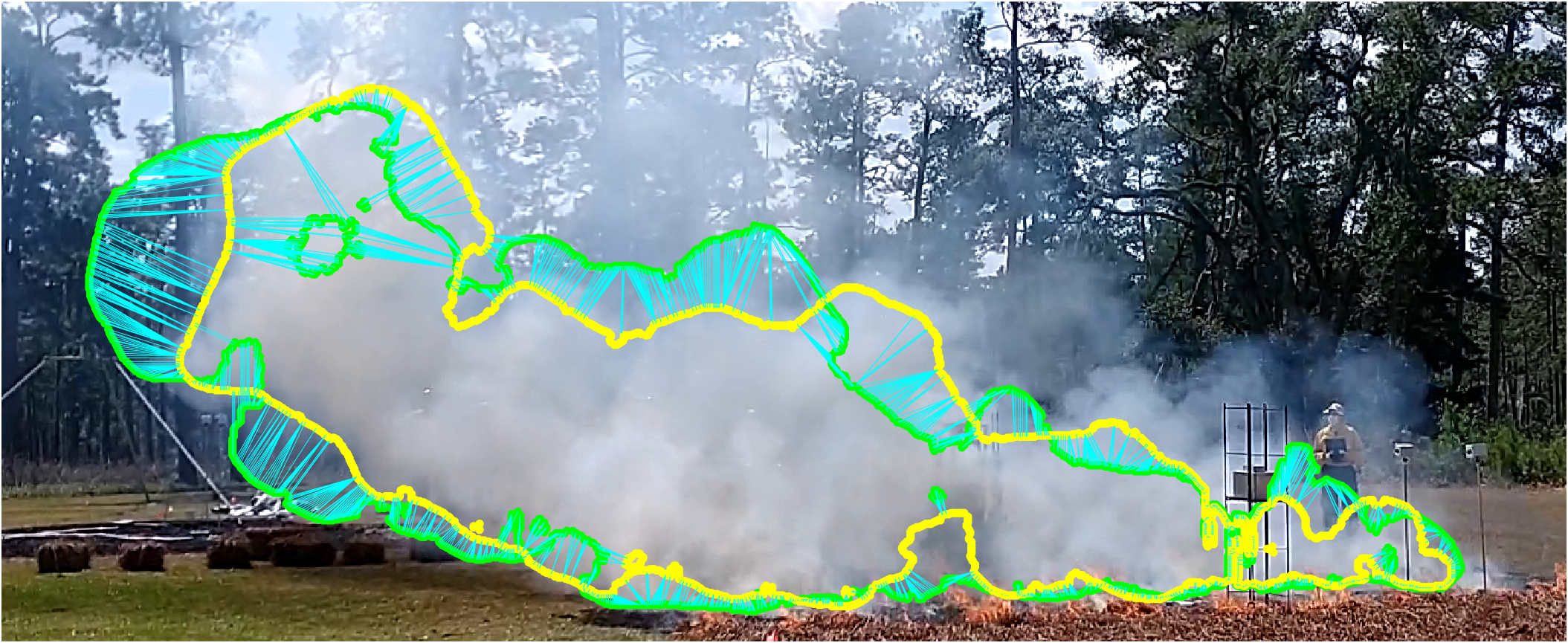}
  \end{center}
  \caption{\label{fig:plumevisualization} {\em Left:} A sample frame
  from a cell phone video of plume spread along a 10~m~$\times$~10~m
  pine straw plot. The superimposed lines show the mean direction of
  spread and are $16^\circ$ from the horizontal. {\em Right:}
  Visualization of displacements calculated between the sample frame and
  successive frame at 1~Hz using FDV. The current frame's boundary is in
  yellow, the successive frame's boundary is in green, and calculated
  displacements between the two are shown with blue lines.}
\end{figure*}

To showcase the application of FDV to near-field plume boundary
evolution in the $x$-$z$ plane, we analyze footage from a
10~m~$\times$~10~m pine straw plot burn conducted by the Environmental
Protection Agency at Tall Timbers Research Station in February 2021. We
recorded the video at 30~Hz using a 2020 Motorola Moto G Power, a
standard low-cost cell phone with dimensions $1920~\times~1080$ pixels.
The field of view and resolution are $\mathtt{FOV} = 1520$~px and
$\mathtt{RES} = 1.10$~px/cm, respectively. $\mathtt{FOV}$ is taken along
the horizontal axis, which is the primary direction of spread in this
video. A sample frame is shown in Figure~\ref{fig:plumevisualization},
alongside a visualization of calculated boundaries and displacements. A
complete set of visualized results for all visual plume frames sampled
at 10~Hz is provided in the
\href{https://github.com/quaife/FDV_Data}{Supplementary Material}.

\begin{figure*}[h]
\begin{center}
\includegraphics[width=0.45\linewidth]{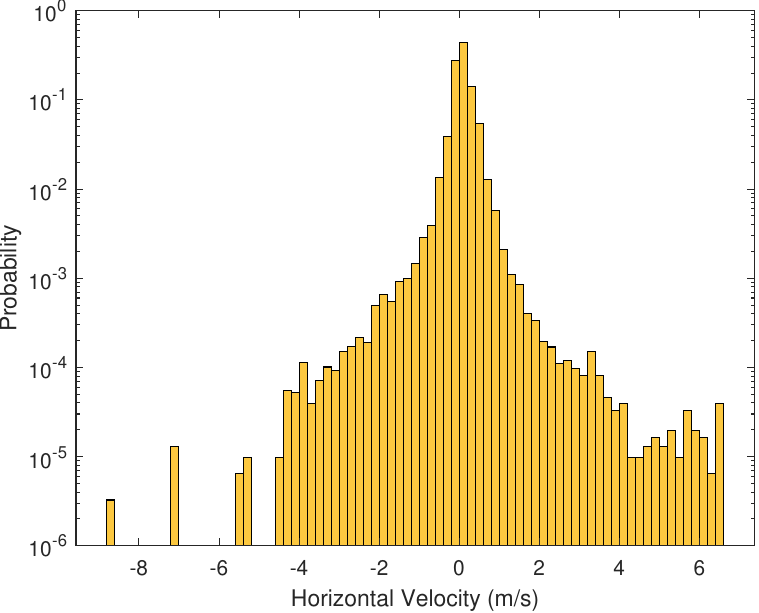}
\quad
\includegraphics[width=0.45\linewidth]{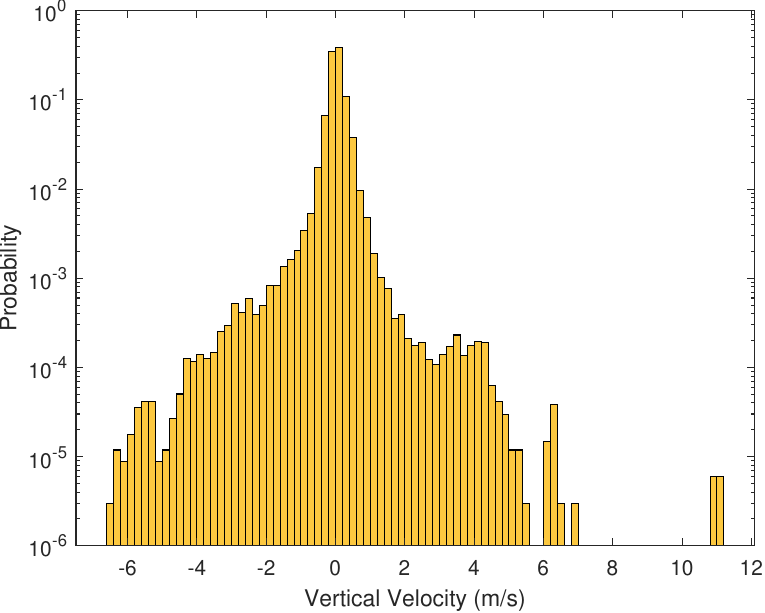}
\end{center}
   \caption{{\em Left:} The horizontal velocity distribution in semi-log
   scale with mean $\mu_H=7.60 \times 10^{-2}$~m/s and standard
   deviation $\sigma_H=0.31$~m/s. The positive direction is from right
   to left in Figure~\ref{fig:plumevisualization}. {\em Right:} The
   vertical velocity distribution in semi-log scale with mean
   $\mu_V=2.23 \times 10^{-2}$~m/s and standard deviation
   $\sigma_V=0.37$~m/s.}
\label{fig:plume}
\end{figure*}

Using the relationship between maximum measurable velocity and sampling
frequency defined in Section~\ref{sec:samplingrate}, we determine that
\begin{align*}
  u_{\max} = \frac{f}{2} \times \frac{\mathtt{FOV}}{\mathtt{RES}} = 8.36f,
\end{align*}
where $f$ is measured in Hz and $u_{\max}$ is measured in m/s, a
customary unit for describing wind speed. Once again, we report the
maximum measurable velocity ($u_{\max}$) and maximum observed velocity
($u_{\mathrm{obs}}$) for several sampling rates ($f$) in
Table~\ref{tab:shannon2}. We observe a clear trend where, as $f$
increases from 1~Hz to 10~Hz, $u_{\mathrm{obs}}$ remains nearly
constant. Beyond 10~Hz, $u_{\mathrm{obs}}$ scales exactly with the
sampling rate and remains at a constant ratio. This strongly indicates
that results from frames sampled above 10~Hz contain values dictated by
the sampling rate rather than by the displacements. Additionally, 10~Hz
is a common sampling rate for wind speed and wind-driven behavior in
fire environments~\cite{heilman2021turbulent, zhou2016effect}. As such,
Figure~\ref{fig:plume} presents horizontal and vertical velocity results
for this experiment from the 10~Hz sampled data.

The process of determining a sampling frequency or maximum measurable
velocity implemented here and in Section~\ref{sec:samplingrate} may be
used for other videos. The ``ideal'' sampling frequencies determined in
this paper are specific to the information captured in these
experiments, and are subject to change based on differences in video
resolution, physical scale relative to the image size, and choice of
entity to track.

\begin{table}[htp]
  \begin{center}
  {\small{
  \begin{tabular}{cccc}
    \toprule
$f$ (Hz)  & $u_{\max}$ (m/s) & $u_{\mathrm{obs}}$ (m/s) & Ratio \\
\midrule
1  & 8.36   & 6.30  & 75\% \\
2  & 16.72  & 5.70  & 34\% \\
5  & 41.80  & 6.07  & 15\% \\
10 & 83.60  & 6.60  &  8\% \\
15 & 125.40 & 18.07 & 14\% \\
30 & 250.80 & 36.14 & 14\% \\
\bottomrule
  \end{tabular}
  }}
  \end{center}
  \caption{\label{tab:shannon2} The maximum measurable velocity ($u_{\max}$) and maximum observed velocity ($u_{\mathrm{obs}}$) at six sampling rates. Also reported is the ratio between the observed velocities and the maximum measurable velocities.}
\end{table}

The horizontal velocities are oriented such that the positive direction
is to the left, aligning with the main direction of flow along that
axis. These results do not fit an obvious classical distribution, unlike
the fire spread results in Sections~\ref{sec:MM} and~\ref{sec:small_xy},
so we do not include any fits calculated with MCMC. The horizontal mean
velocity, $\mu_H=7.60 \times 10^{-2}$~m/s, and vertical mean velocity,
$\mu_V=2.23 \times 10^{-2}$~m/s, are both positive, which agrees with
the net leftward and upward motion of the plume. Moreover, the
ratio of the two mean values results in an inclination angle of the
plume from the ground of approximately $16^\circ$. Lines with these
slopes are superimposed in Figure~\ref{fig:plumevisualization}, and they
visually agree well with the general direction of spread of the
near-field plume.

The mean value of the horizontal velocities is very close to zero, which
might seem contradictory to the predominantly leftward motion. However,
high temporal and spatial resolutions allow for the measurement of
fine-scale displacements. These displacements are often the
concentration or dissipation of smoke, rather than a net advective
movement, and the likelihood of such displacements being positive or
negative is approximately equal. This results in the large amount of
symmetry seen in the horizontal velocity distribution. However, there is
some mass in the distribution at velocities greater than 4~m/s, and
these samples correspond to the general downwind motion of the plume as
it fills in gaps in the smoke column. Decreasing the sampling rate
reduces the number of small-scale motions, resulting in larger mean
values for the horizontal velocities. In particular, as shown in
Table~\ref{tab:meanHorizontalVelocities}, we observe that lower sampling
rates result in larger mean horizontal velocities.

\begin{table}
  \centering
  \begin{tabular}{c|ccccc}
    $f$ (Hz) & 1 & 2 & 3 & 5 & 10 \\
    \hline
    $\mu_H$ (m/s)   & $1.75 \times 10^{-1}$ & $1.38 \times 10^{-1}$
                    & $1.19 \times 10^{-1}$ & $9.84 \times 10^{-2}$ 
                    & $7.60 \times 10^{-2}$ \\
    $\sigma_H$ (m/s) & $9.90 \times 10^{-1}$ & $6.92 \times 10^{-1}$
                    & $5.29 \times 10^{-1}$ & $4.32 \times 10^{-1}$ 
                    & $3.10 \times 10^{-1}$
  \end{tabular}
  \caption{\label{tab:meanHorizontalVelocities} The mean ($\mu_H$) and standard deviation ($\sigma_H$) of the horizontal velocity for various sampling frequencies, $f$. As with the ratio between $u_{\max}$ and $u_{\mathrm{obs}}$ in Table~\ref{tab:shannon2}, both values decrease as $f$ increases.}
\end{table}

It is worth noting that the standard deviation of the horizontal and
vertical velocities are significantly larger than their corresponding
mean values. In the horizontal direction, the standard deviation is
about four times larger than the mean value, and in the vertical
direction, the ratio is more than an order of magnitude larger.  Such
large variances differ from our statistical analysis of a moving fire
front (Section~\ref{sec:small_xy}), but this is to be expected since
turbulent wind behavior has a more lasting effect on the plume's motions
than on the flames'. Additionally, smoke dispersion leads to higher
levels of uncertainty; the plume dissipates upon reaching areas of lower
concentration, causing sections of smoke to no longer be visible. FDV
interprets these changes as displacements toward the remaining main body
of the plume. This is similar in cause and effect to the flame
flickering behavior discussed in Section~\ref{sec:IRandVIS}. Plume
segmentation also introduces more uncertainty than fire segmentation
since the visual appearance of plumes vary significantly based on the
smoke concentration at any given point, whereas flames have similar
bright colors regardless of the amount of flame present.

%%%%%%%%%%%%%%%
\subsection{Comparison with Other Methodologies}
\label{sec:compare}

In this section, we compare FDV's segmentation and velocity calculations
for the visual fire (Section~\ref{sec:small_xy}) and visual plume
(Section~\ref{sec:xz}) videos with those of other computer vision
methods. FDV applies the same processing steps to both infrared and
visual images following segmentation. Therefore, since infrared image
segmentation in FDV depends on physical temperature ranges rather than
algorithm selection, we exclude infrared images from this comparison.

\subsubsection{Image Segmentation}

Segmentation approaches in fire science and other applications commonly
rely on thresholding with one or more color spaces
(Section~\ref{sec:relatedseg}). Another widely used approach in computer
vision is $k$-means cluster image segmentation~\cite{shan2018image},
which groups pixels into clusters based on their attributes. While
adaptive $k$-means segmentation
methods~\cite{zheng2018image,sulaiman2010adaptive} are available, these
require significantly more computation than FDV's thresholding approach.
As such, we use standard $k$-means segmentation for comparison.

\begin{figure*}[h]
\begin{center}
\includegraphics[width=0.3\linewidth]{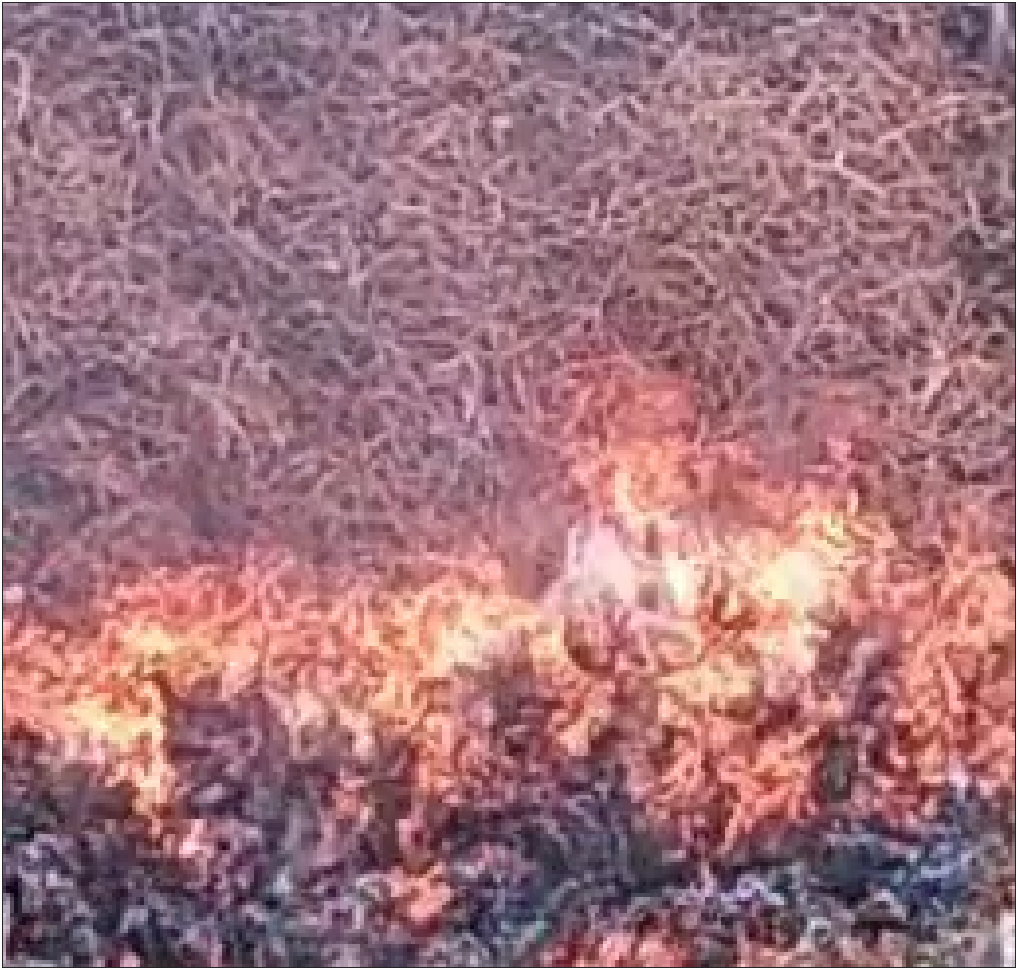}
\quad
\includegraphics[width=0.299\linewidth]{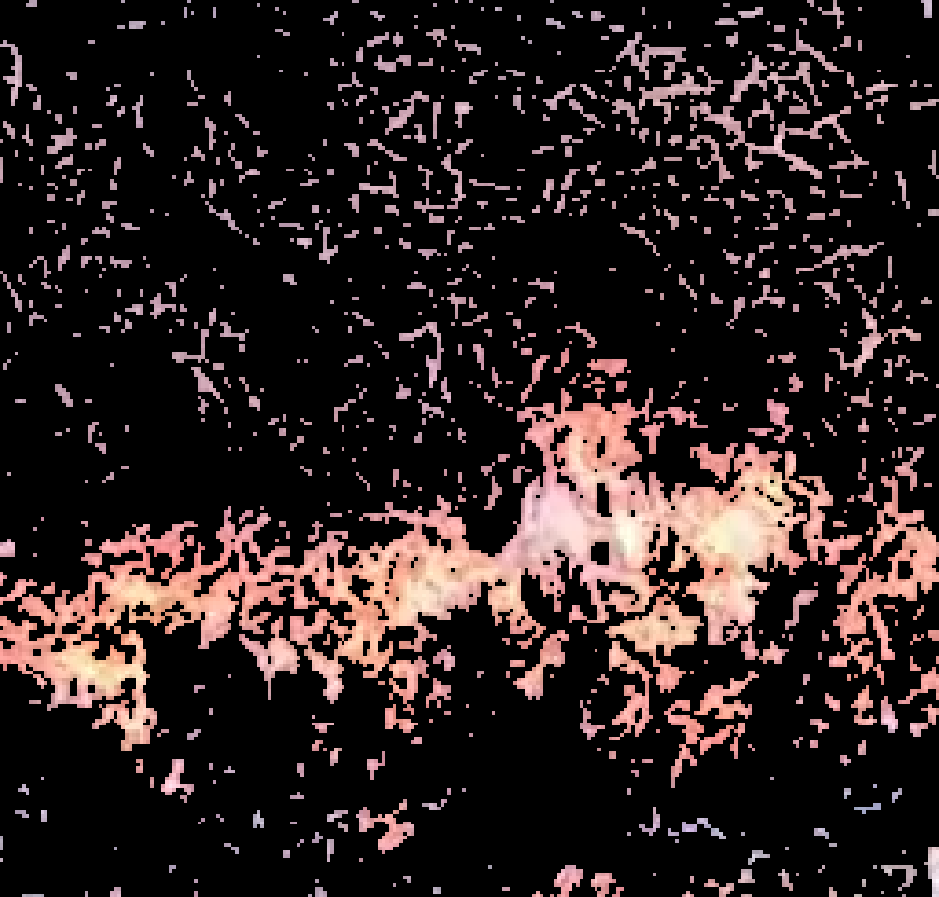}
\quad
\includegraphics[width=0.3\linewidth]{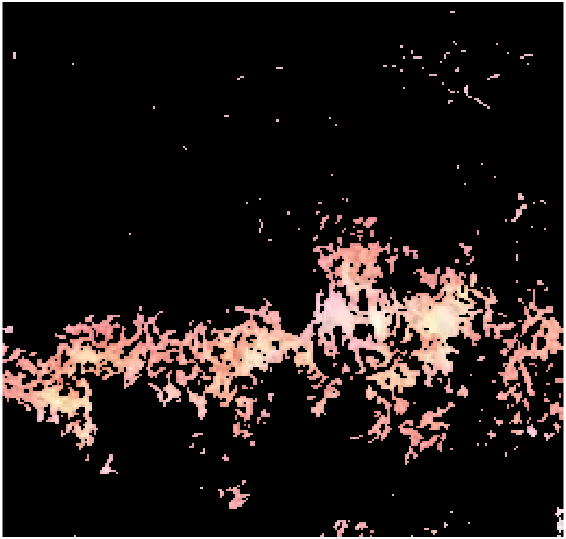}
\end{center}
  \caption{{\em Left:} The original image. {\em Middle:} The $k$-means
  cluster segmentation in OpenCV with $k=3$ clusters. {\em Right:} FDV's
  combined RGB and HSV segmentation. No cleaning has been performed on
  the segmented image.}
\label{fig:KMEANSfire}
\end{figure*}

Figure~\ref{fig:KMEANSfire} compares the results of segmenting a sample
fire image using $k$-means cluster segmentation in
OpenCV~\cite{bradski2008learning} and RGB-HSV thresholding in FDV. Both
methods successfully identify flames in the image, but FDV's combined
color space thresholding more effectively distinguishes flames from the
surrounding fuel. Additionally, $k$-means segmentation is more
computationally expensive than basic thresholding. Another common
segmentation method, the watershed
algorithm~\cite{kornilov2018overview}, proved unreliable for small-scale
fire image segmentation due to substantial brightness variations across
flame regions.

\begin{figure*}[h]
\begin{center}
\includegraphics[width=0.3\linewidth]{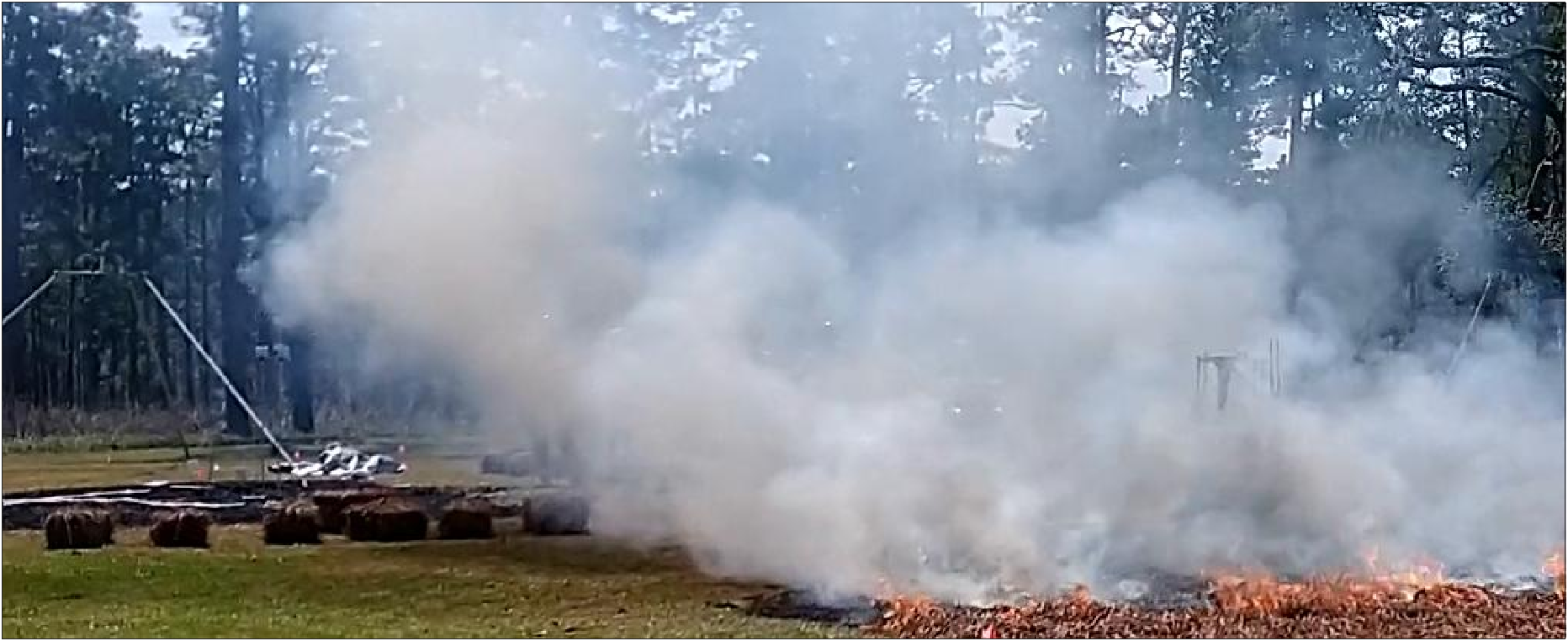}
\quad
\includegraphics[width=0.3\linewidth]{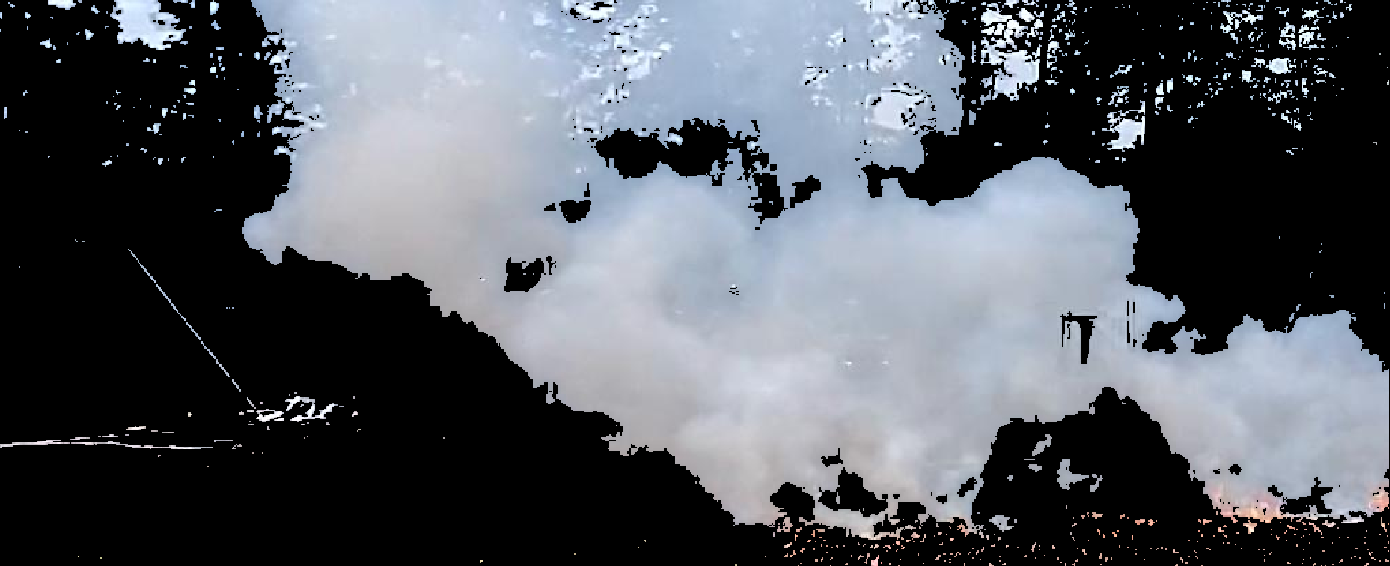}
\quad
\includegraphics[width=0.3\linewidth]{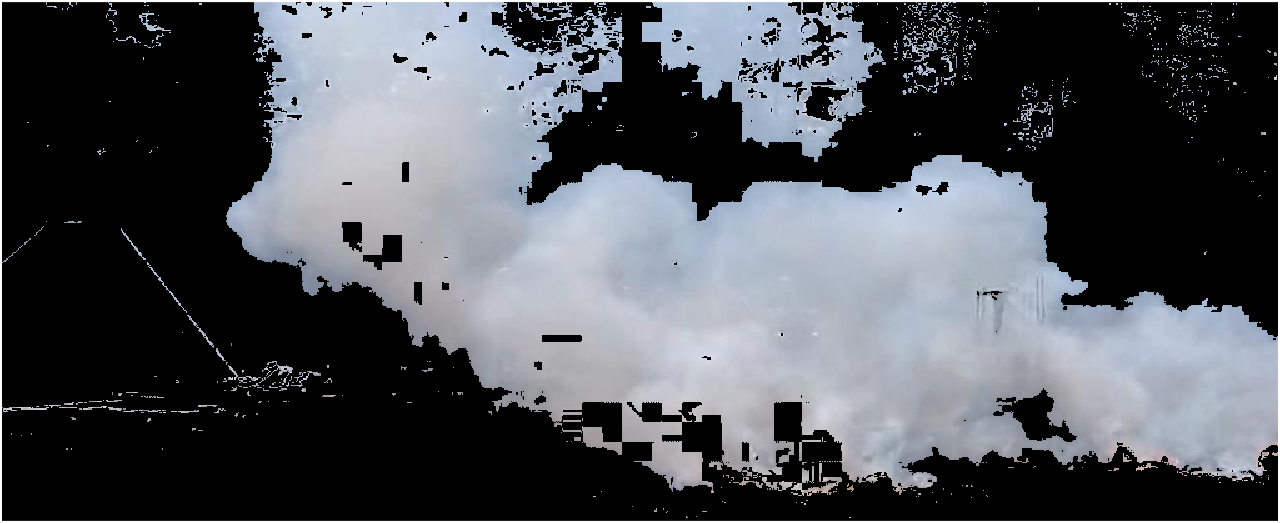}
\end{center}
   \caption{{\em Left:} The original image. {\em Middle:} The
   $k$-means cluster segmentation in OpenCV with $k=4$ clusters. {\em Right:} FDV's combined RGB and HSV segmentation. No cleaning has been performed on the segmented image.}
\label{fig:KMEANSplume}
\end{figure*}

Figure~\ref{fig:KMEANSplume} compares the output of $k$-means
segmentation and RGB-HSV thresholding in FDV for the plume.  Increasing
the number of $k$-means clusters removes significant portions of the
plume, whereas decreasing the number of clusters retains excess area
around it. Although both segmentation methods capture points outside of
the plume, FDV's approach fully excludes the fire and some areas of sky
that $k$-means includes. While carefully tuning the $k$-means parameters
or adopting variants of the algorithm may improve segmentation, FDV
achieves effective results without this additional complexity.

\subsubsection{Velocity Calculation}

As stated in Section~\ref{sec:relatedtrack}, PIV is a common choice for
velocity calculations in fire dynamics, stemming from its widespread
success in fluid dynamics research. We use
PIVlab~\cite{stamhuis2014pivlab,thielicke2021particle}, the preeminent
PIV software in MATLAB, as a basis of comparison for the velocities
calculated by FDV.

\begin{figure*}[h]
\begin{center}
\includegraphics[width=0.45\linewidth]{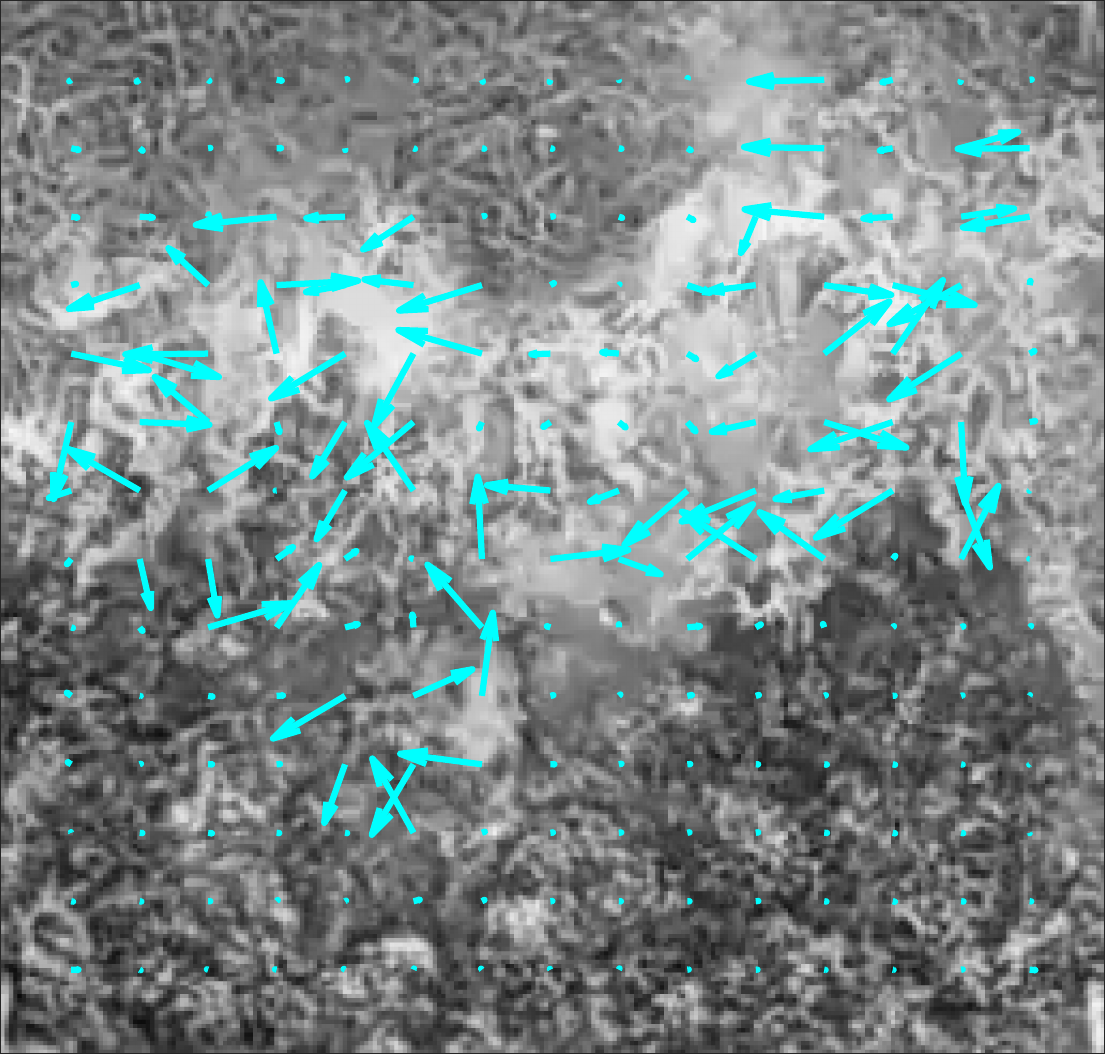}
\quad
\includegraphics[width=0.45\linewidth]{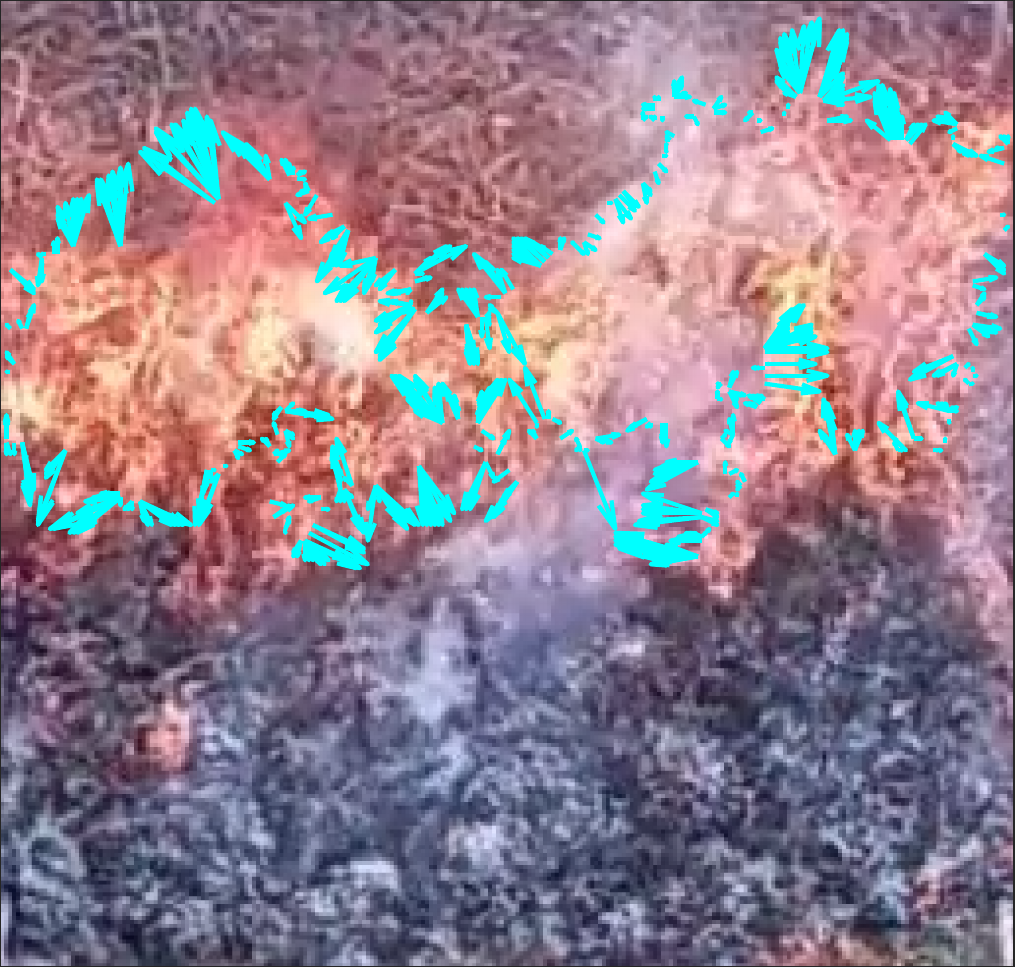}
\end{center}
   \caption{{\em Left:} A sample frame of output from PIVlab using the
   software's suggested parameter values, visualized with blue velocity
   vectors. {\em Right:} The same sample frame with FDV's output,
   visualized using blue velocity vectors.}
\label{fig:PIVfire}
\end{figure*}

Figure~\ref{fig:PIVfire} visualizes the velocities calculated by PIVlab
and FDV for the same frame pair in the visual fire video. The
displacements calculated with PIV are primarily unphysical across all
frames, despite using the suggested parameter values. Many vectors cross
each other, track the smoke instead of the fire, or fail to match the
actual motions observed. Optical flow methods, including PIV, generally
rely on assumptions that the appearance of objects or points of interest
remain nearly constant across frames. The changing brightness and
turbulence of flames violate these assumptions, whereas FDV is
specifically designed to accommodate the unique visual characteristics
and behavior of flames.

It is important to note that the errors associated with PIV results are
not due to deficiencies within PIVlab; we observe similar errors for
other PIV software and optical flow techniques. Though the accuracy of
PIV calculations might improve with careful parameter tuning or
additional pre-processing, FDV does not share this sensitivity to
parameter choices. PIV yields more reasonable results in fire
experiments when the flow is seeded with
particles~\cite{desai2022investigating}, but FDV does not require a
similar experimental setup.

\begin{figure*}[h]
\begin{center}
\includegraphics[width=0.45\linewidth]{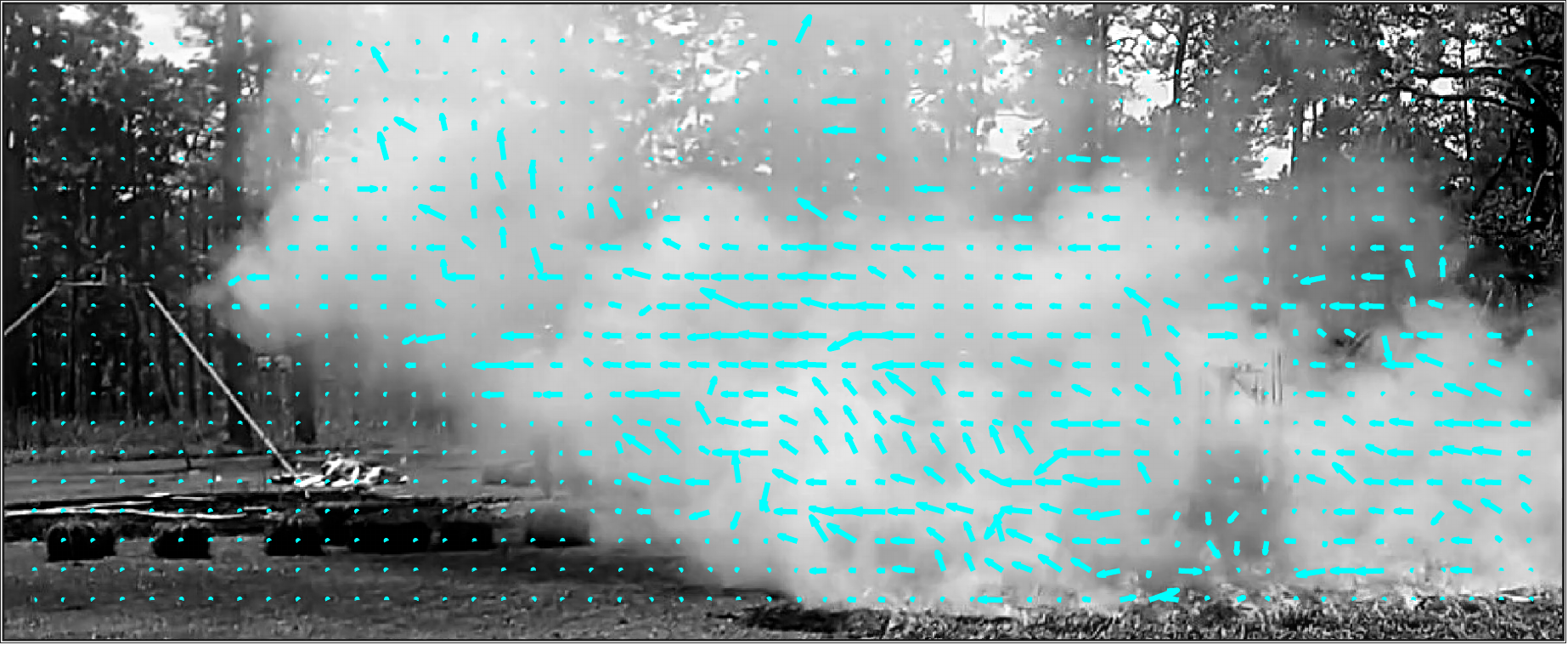}
\quad
\includegraphics[width=0.45\linewidth]{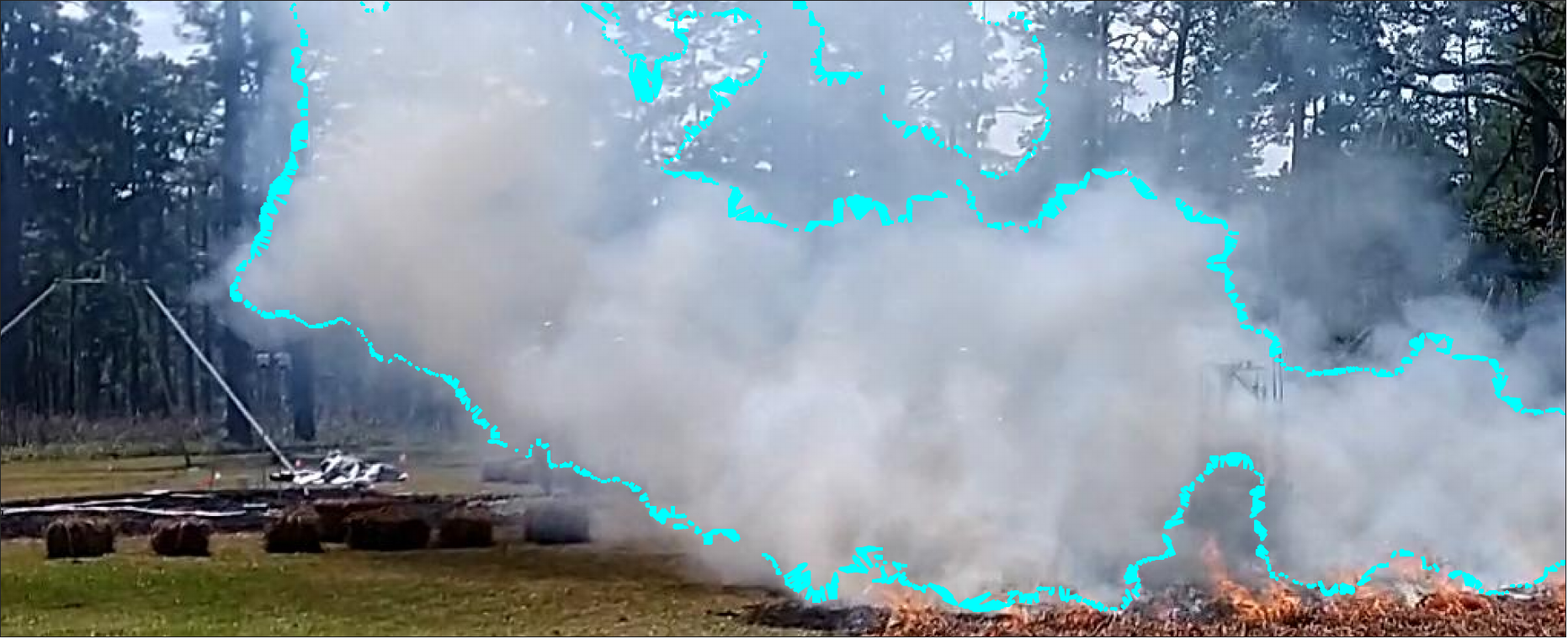}
\end{center}
   \caption{{\em Left:} A sample frame of output from PIVlab using the
   software's suggested parameter values, visualized with blue velocity
   vectors. {\em Right:} The same sample frame with FDV's output,
   visualized using blue velocity vectors.}
\label{fig:PIVplume}
\end{figure*}

Figure~\ref{fig:PIVplume} visualizes the velocities calculated by PIVlab
and FDV for the same frame pair in the visual plume video. We see that
the plume is a more suitable candidate for PIV than fire, as it has
greater visual consistency between frames. In this context, PIV can be
an excellent choice for determining velocities within a two-dimensional
domain. However, a direct comparison with FDV is not feasible, as FDV
tracks the displacement of an interface rather than internal and
external velocities---two different quantities relative to the plume's
motion.

%%%%%%%%%%%%%%%
\subsection{Further Applications and Future Work}
\label{sec:future}

\begin{figure*}[h]
\begin{center}
\includegraphics[width=0.4\linewidth,trim=15.5cm 3cm 7cm 0.5cm,clip=true]{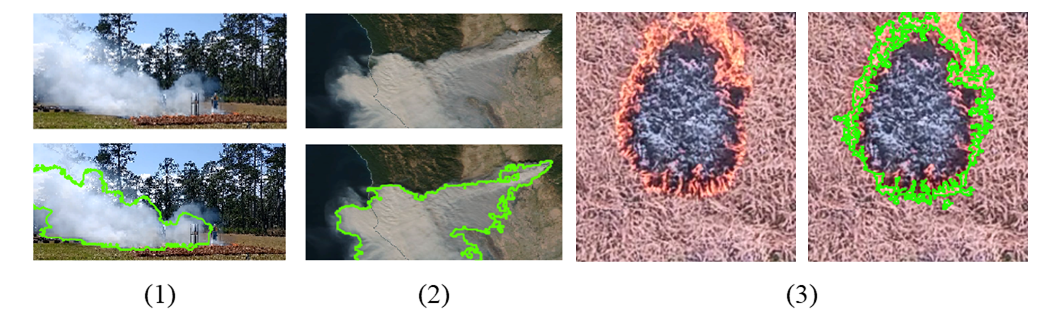}
\hspace{0.1cm}
\includegraphics[width=0.4\linewidth,trim=21.5cm 3cm 1cm 0.5cm,clip=true]{boundaries.png}

\vspace{0.3cm}

\includegraphics[width=0.4\linewidth,trim=0.25cm 0.1cm 14.6cm 0.1cm,clip=true]{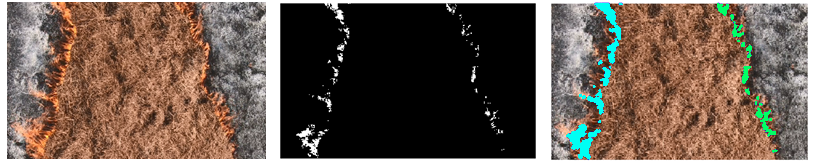}
\hspace{0.1cm}
\includegraphics[width=0.4\linewidth,trim=14.6cm 0.1cm 0.25cm 0.1cm,clip=true]{multi2.png}
\end{center}
  \caption{Sample fire segmentation and boundary calculation results for
  a sub-microscale ring fire ({\em top}) and flanking fire fronts ({\em
  bottom}) captured in the $x$-$y$ plane. Ring fire flames are outlined
  in green and flanking fire flames are denoted in blue and green to
  differentiate the flanking regions. Both experiments are performed on
  pine straw plots and recorded with a Hero 7 GoPro.}
\label{fig:bound}
\end{figure*}

To expand on the primarily unidirectional fire spread scenarios
described in Sections~\ref{sec:small_xy}~and~\ref{sec:xz}, we showcase
the results of segmentation and boundary calculation for two additional
fire geometries; a ring fire and two approaching flanking fire fronts,
displayed in Figure~\ref{fig:bound}. 
The flanking fronts example, in particular, demonstrates FDV's
ability to segment multiple regions. In this application, the two flanks are separated using DBSCAN parameters $\epsilon = 20$ and $\mathtt{minPts} = 10$ for an image size of $163 \times 268$~px. Proper analysis of multidirectional spread and the interaction between
multiple flame fronts require additional analysis that we reserve for
future work.

\begin{figure*}[h]
\begin{center}
\includegraphics[width=0.22\linewidth]{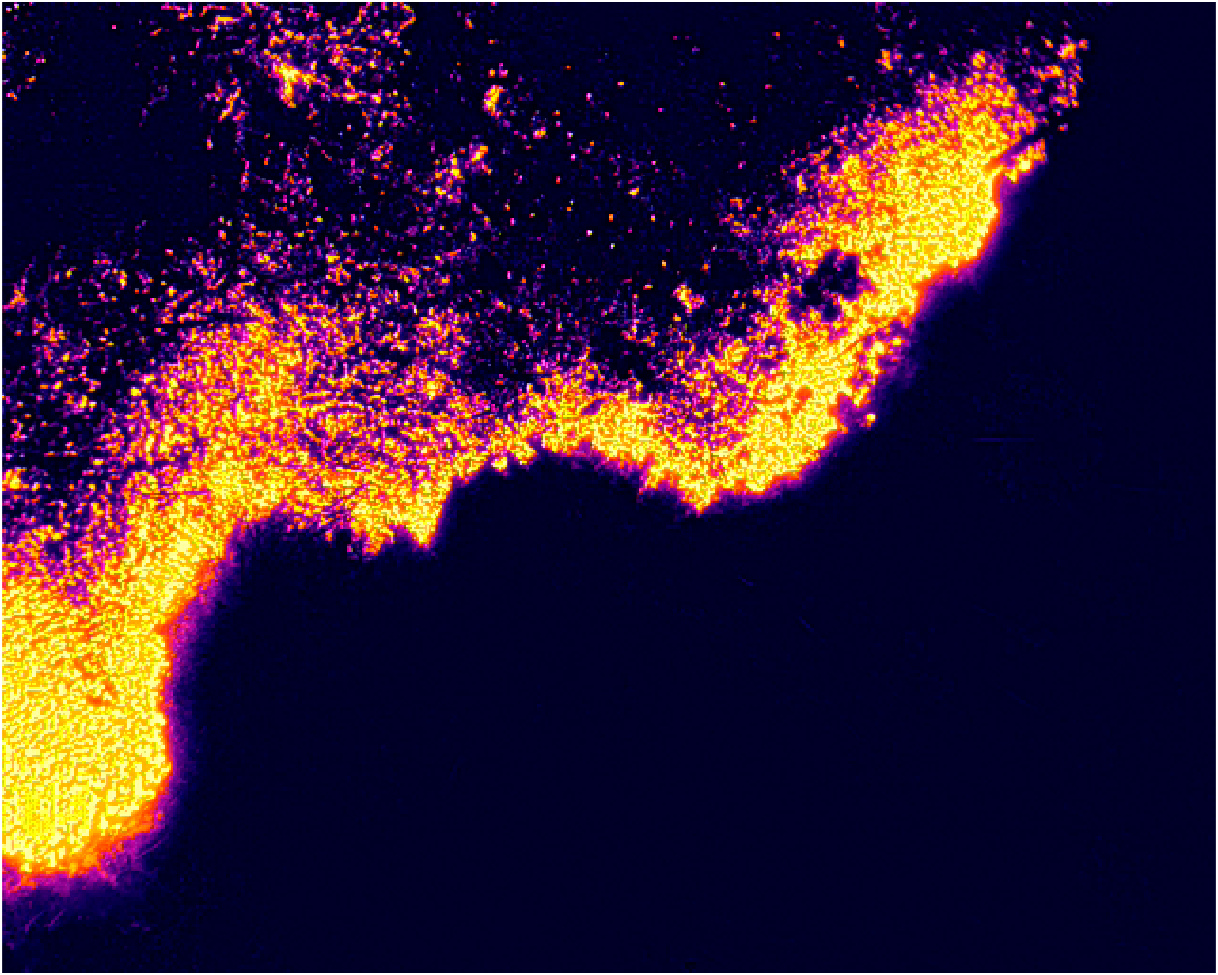}
\includegraphics[width=0.22\linewidth]{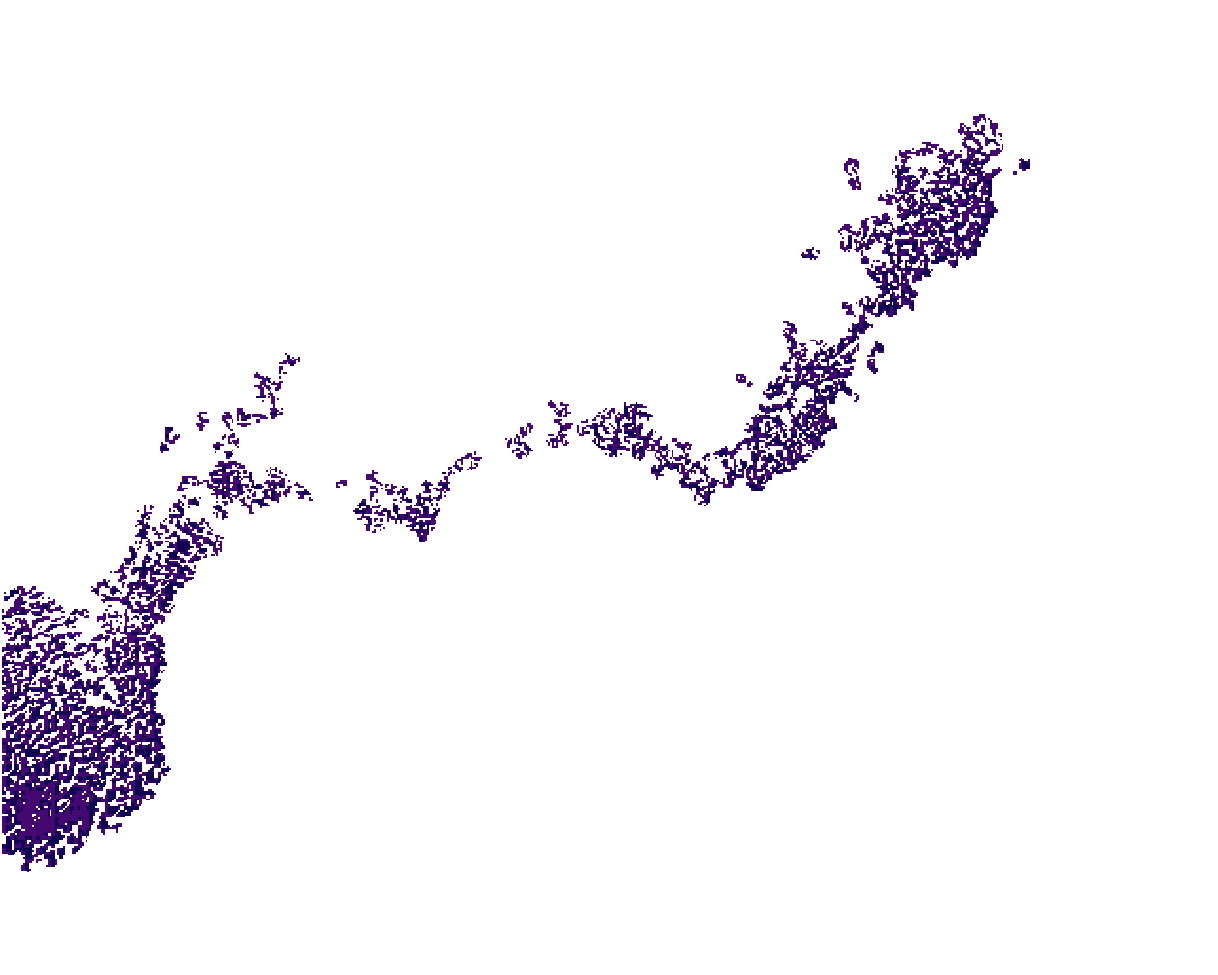}
\includegraphics[width=0.22\linewidth]{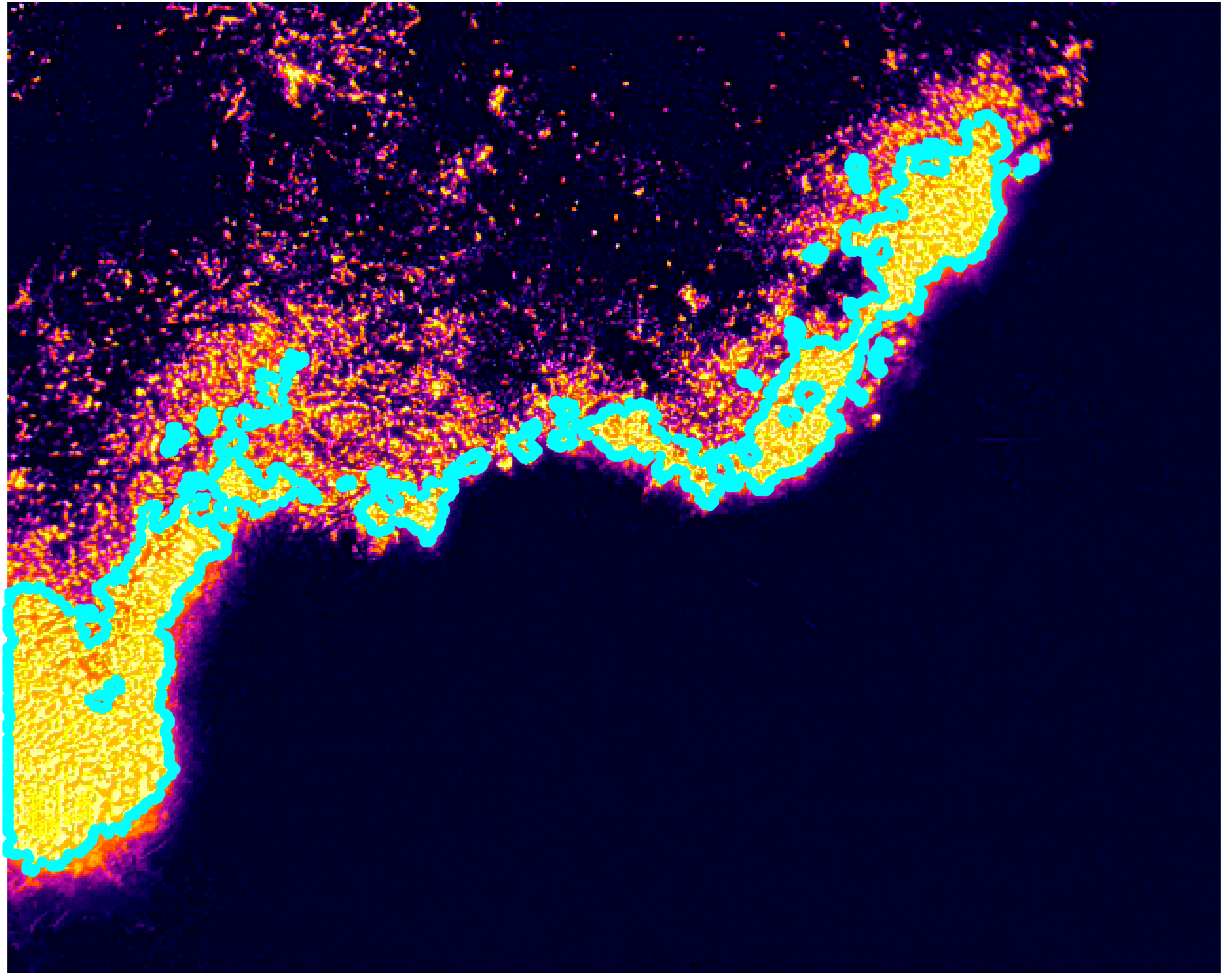}
\includegraphics[width=0.22\linewidth]{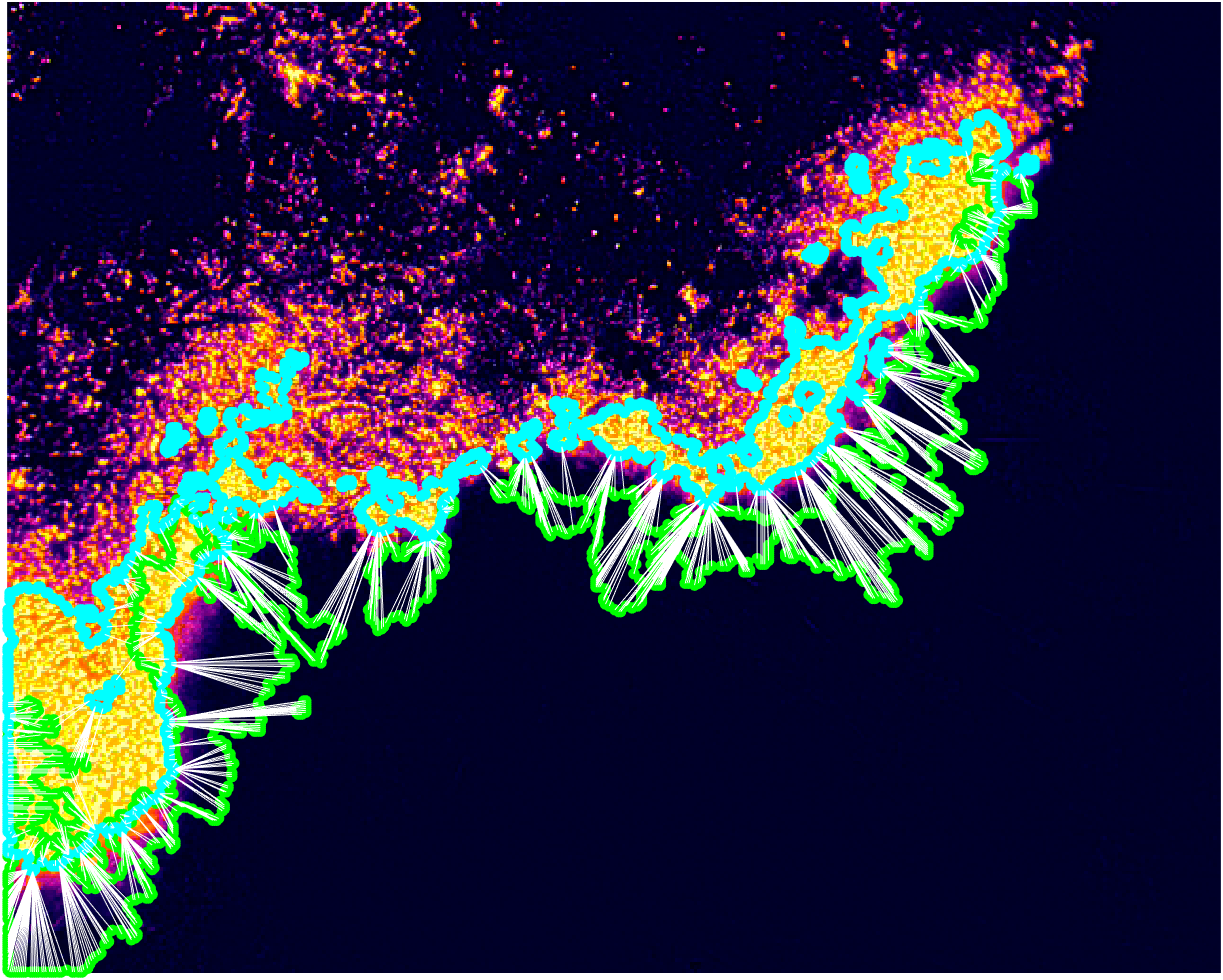}
\end{center}
   \caption{Visualization of a segment of the FDV pipeline applied to processing microscale observations obtained via UAV. From left to right, the figure shows: the original infrared image, the temperature-segmented image, the boundary of the burning region, and the displacement between this boundary and the boundary in the next timestep.}
\label{fig:xydrone}
\end{figure*}

The principles implemented to analyze infrared and visual image
sequences in Section~\ref{sec:small_xy} apply directly to other infrared
and visual image datasets. Figure~\ref{fig:xydrone} showcases a midwave
infrared image from one such dataset---microscale $x$-$y$ fire spread
captured via UAV. In this example, FDV segments and cleans the burning
area within the infrared image, calculates its boundary, and determines
its displacement based on the subsequent timestep. Using images from
UAVs or other non-stationary data capture frameworks, however,
necessitate a discussion of image stabilization. Image stabilization
methods differ depending on several factors, most notably whether the
images contain associated geospatial coordinate data or
markers~\cite{paugam2021orthorectification}. Fire environments present
unique challenges for this problem due to constant motion from most
elements in the frame and the variable flame visibility discussed in
Section~\ref{sec:IRandVIS}. Our investigation into optimal algorithmic
choices for image stabilization and quantifying the error introduced by
each method is ongoing. As such, detailed calculations from videos
obtained with a UAV is forthcoming.

\begin{figure*}[h]
\begin{center}
\includegraphics[width=0.45\linewidth]{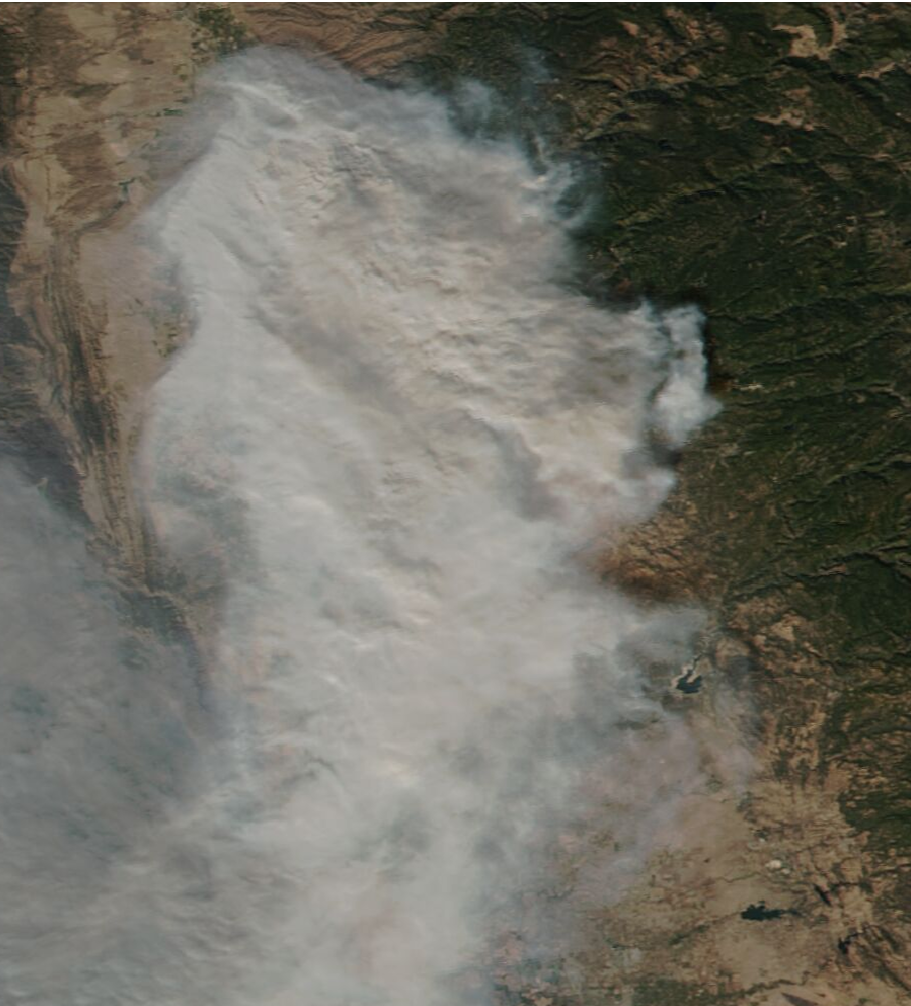}
\quad
\includegraphics[width=0.45\linewidth]{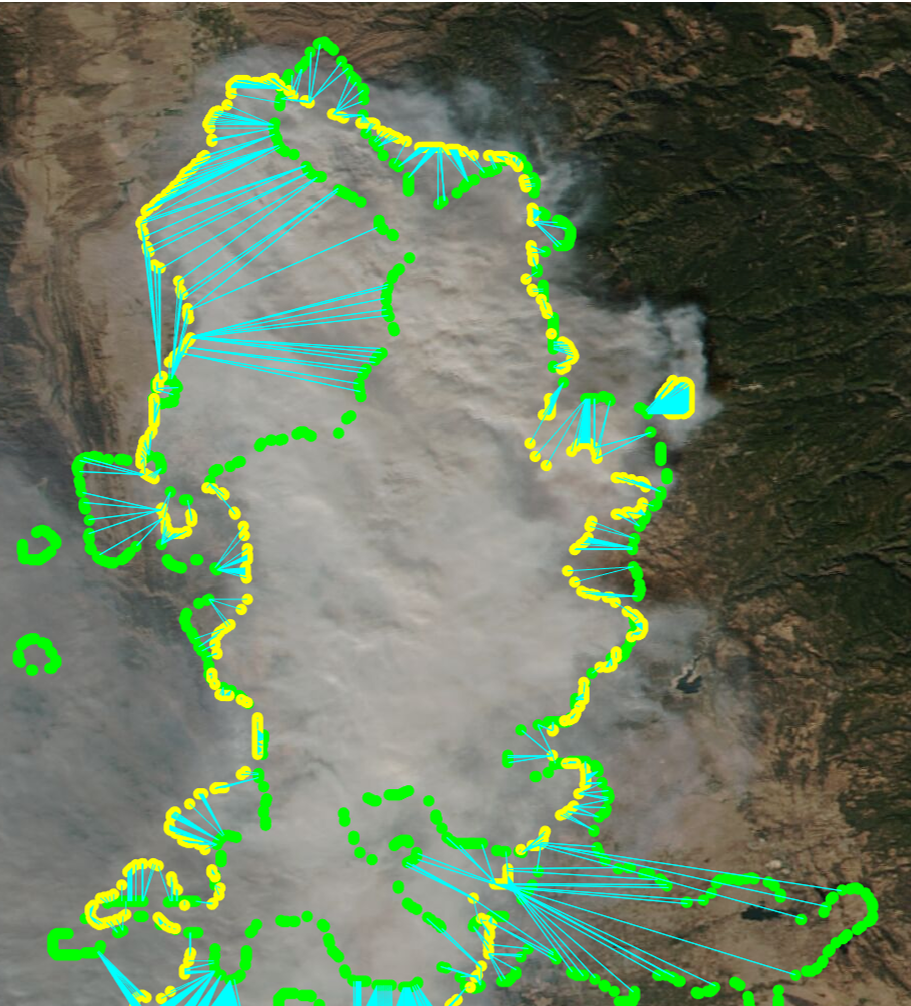}
\end{center}
   \caption{\label{fig:sat} {\em Left:} Satellite image of the Camp Fire
   in Butte County, CA on November 6, 2018. These data are downloaded
   from the \href{https://worldview.earthdata.nasa.gov}{NASA Worldview
   Application}, part of the NASA Earth Observing System Data and
   Information System (EOSDIS)~\cite{unknown-author-no-date}. This image
   comes from the VIIRS corrected reflectance imagery located in NASA's
   Wildfires Data Pathfinder system. {\em Right:} Sample displacements
   (blue) between the boundary of the plume as pictured (yellow) and the
   boundary of the plume approximately one day later (green).}
\end{figure*}

Moving farther out in spatial scale, we showcase the application of FDV to satellite imagery and its ability to analyze synoptic-scale observations. The sample satellite image in Figure~\ref{fig:sat} is obtained from the \href{https://worldview.earthdata.nasa.gov}{NASA Worldview Application} and has a per-pixel spatial resolution of approximately 500~m. As this is meant to simply demonstrate the capabilities of FDV, we leave a further analysis of synoptic-scale plume dynamics for future work. It is worth mentioning that FDV performs in the same way for other satellite-based measurements and observations related to fire research, such as concentrations of atmospheric chemical compounds for smoke pollution studies.

Plans for other applications of FDV combine aspects from the examples
given throughout Section~\ref{sec:experiments}. Possible directions
include plume analysis
using combined visual and infrared data and extended plume physics
calculations involving vorticity, sweeps and ejections, and more diverse
plume scenarios. Preliminary analysis has revealed similarity in trends
calculated between plume experiments, suggesting that the results
obtained are not strictly scenario-specific and may extend beyond the
exact set of initial conditions for a given experiment.

%% ---------------------------
%% Conclusion
%% ---------------------------
\section{Conclusion} %% {{{
\label{sec:conclusion}

FDV is a comprehensive tool designed to process both infrared and visual videos of fire and plume behavior that accommodates various angles, spatial scales, and image capture frameworks. Whether sourced from a cell phone, FLIR infrared camera, visual GoPro, satellite, or other devices, FDV identifies boundaries for regions of interest and calculates velocities at these boundary points, offering detailed and dynamic insights into fire and plume behavior. In addition to its analytical capabilities, FDV automatically generates spatiotemporally-formatted datasets for specified regions and calculations of interest. These datasets streamline the integration of observational data into predictive and analytical models. This automation enhances the efficiency and accuracy of fire behavior analysis, making FDV a valuable tool for researchers and practitioners in the field.

We demonstrate that MCMC provides a more accurate fit for classical distributions compared to moment matching, thus improving the statistical analyses within FDV. We apply the updated FDV algorithm to the same infrared video dataset used in our previous methodology~\cite{sagel2021fine}, confirming that the results are consistent and comparable in the face of FDV's algorithmic improvements and expansions. Additionally, we compare the outputs of applying FDV to both infrared and visual videos sampled at 1~Hz. The visual data reveal larger positive and more negative velocities, reflecting flame dynamics that the infrared data do not capture.

Furthermore, we determine optimal sampling frequencies using
the Nyquist-Shannon sampling theorem. This approach yields sampling frequencies that agree with values from literature. We observe
that sampling a visual video at 2~Hz, rather than 1~Hz, produces
velocity distributions with similar overall shapes but larger
velocities, as expected. The process of determining ideal sampling
frequency is also applied to a visual plume video, and again yields a
suggested value that is consistent with common experimental choices and
literature. The statistics output by FDV from this plume video have
physical interpretations that match the observed behavior.

\subsection*{CRediT author statement}
\medskip
\noindent
{\bf Daryn Sagel:} Conceptualization, Methodology, Software, Validation,
Formal Analysis, Investigation, Data Curation, Writing - Original Draft,
Writing - Review \& Editing, Visualization. {\bf Bryan Quaife:}
Conceptualization, Methodology, Formal Analysis, Investigation,
Resources, Writing - Review \& Editing, Supervision, Project
Administration, Funding Acquisition.

\subsection*{Data availability}
\medskip
\noindent
Data and code are available at
\href{https://github.com/quaife/FDV_Data}{GitHub}.

\subsection*{Acknowledgments}
\medskip
\noindent
This work is supported by the Department of Defense Strategic
Environmental Research and Development Program (SERDP) RC20-1298.

%% ---------------------------
%% Bibliography and postamble
%% ---------------------------
\bibliographystyle{elsarticle-num}

%\section*{References}

\end{document}